%% file: main.tex
\documentclass[10pt,twocolumn,letterpaper]{article}

\usepackage[pagenumbers]{cvpr} 

\input{preamble}

\definecolor{cvprblue}{rgb}{0.21,0.49,0.74}
\usepackage[pagebackref,breaklinks,colorlinks,allcolors=cvprblue]{hyperref}

\def\paperID{} 
\def\confName{CVPR}
\def\confYear{2026}

\title{PrAda: Few-Shot Visual Adaptation for Text-Prompted Segmentation}

\author{Gabriele Rosi$^{1, 2}$
\quad 
Fabio Cermelli$^{2}$
\quad
Carlo Masone$^{1,2}$
\quad
Barbara Caputo$^{1,2}$\\
$^{1}$ Politecnico di Torino
$^{2}$ Focoos AI \\
{\tt\small \{name.surname\}@polito.it \, \{name.surname\}@focoos.ai}
}

\begin{document}

\maketitle
\input{sec/0_abstract}

\input{sec/1_intro}

\begin{figure*}[t]
  \centering
  \includegraphics[width=1\linewidth]{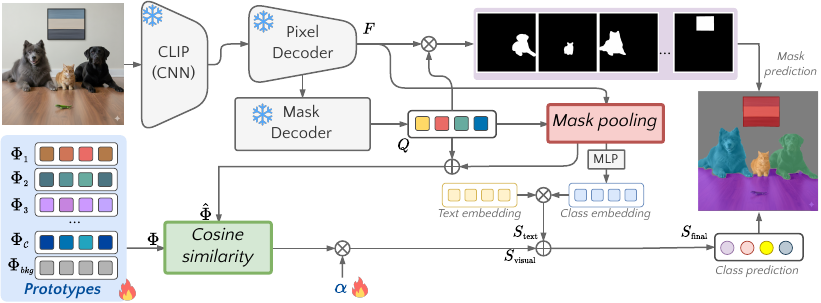}
  \caption{\textbf{PrAda overview}. 
  An input image is passed through a frozen FC-CLIP~\cite{yu2023convolutions} backbone to obtain visual features $F$ and query representations $Q$. The segmentation masks from the decoder are used to pool features, which are then combined with their queries to form representations $\hat{\Phi}$. Cosine similarity with learnable class prototypes $\Phi$, initialized from few-shot exemplars $\mathcal{V}$, gives a visual similarity score $S_{\text{visual}}$. In parallel, text similarity score $S_{\text{text}}$ is computed using the textual embeddings extracted from CLIP~\cite{radford2021learning} text encoder (not shown in figure). The final prediction $S_{\text{final}}$ is computed by fusing both similarities with a learnable parameter $\alpha$.}
  \label{fig:method}
\end{figure*}

\input{sec/2_related}

\begin{figure*}[t]
  \centering
  \includegraphics[width=1\linewidth, height=220px]{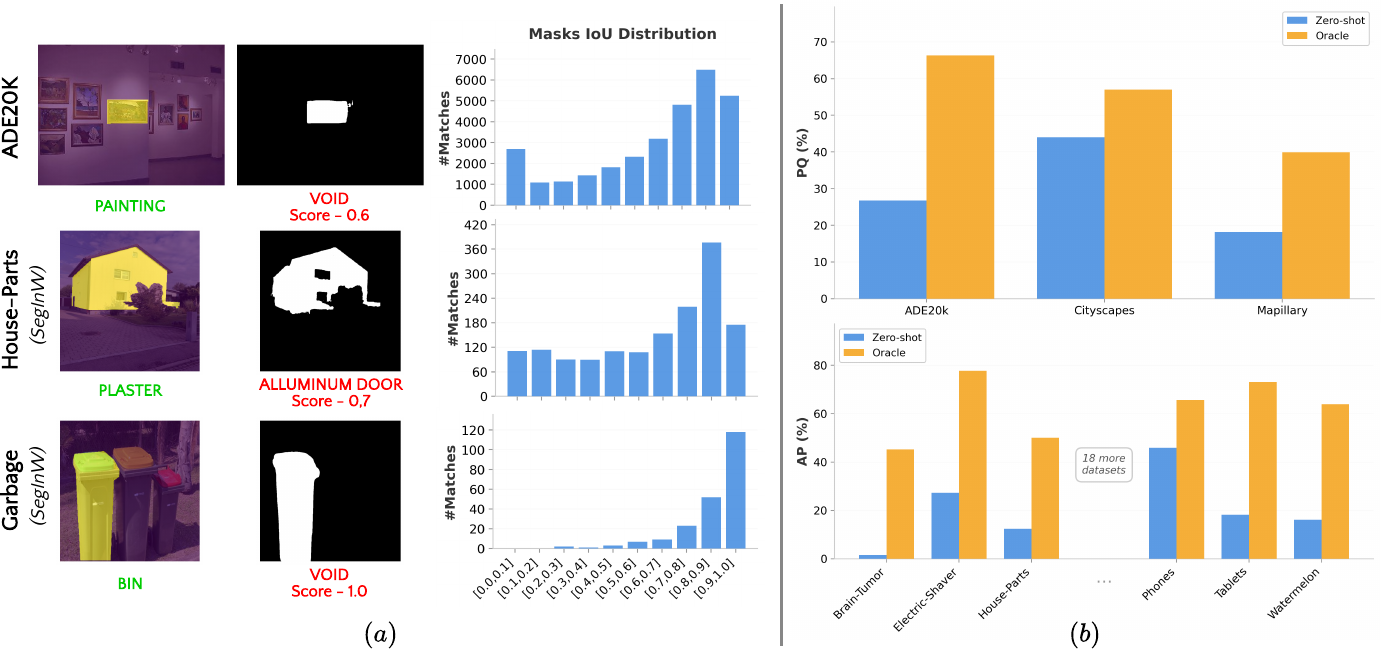}
  \caption{\textbf{Failure mode of text-prompted segmentation.} (a) FC-CLIP~\cite{yu2023convolutions} predictions across different datasets. \textit{Left}: Example of failures, where the model accurately localizes the object mask, but wrongly classifies it. \textit{Right}: Mask IoU distributions across three datasets show that predictions are heavily concentrated in high IoU ranges ([0.8, 1.0]), with the majority of masks achieving IoU $>0.5$ with ground truth. This confirms strong spatial segmentation quality despite misclassification. (b) Performance gap between zero-shot FC-CLIP predictions and the same model with an oracle classifier. The gap across multiple datasets, highlighting that the primary bottleneck is classification rather than mask quality. Datasets in the second row belongs to SegInW \cite{zou2023generalized}.}
  \label{fig:motivations}
\end{figure*}

\input{sec/3_method}
\input{sec/4_results}
\input{sec/5_conclusion}

\newpage
\myparagraph{Acknowledgements}
This publication is part of the project PNRR-NGEU which has received funding from the MUR – DM 117/2023. We acknowledge ISCRA for awarding this project access to the LEONARDO supercomputer, owned by the EuroHPC Joint Undertaking, hosted by CINECA (Italy).

{
  \small
  \bibliographystyle{ieeenat_fullname}
  \bibliography{main}
}

\input{sec/X_suppl}

\end{document}

%% file: preamble.tex
\usepackage{multirow}
\usepackage{colortbl}
\usepackage[table]{xcolor}
\usepackage[accsupp]{axessibility}  

\usepackage{amssymb}
\usepackage{pifont}
\newcommand{\xmark}{\ding{55}}%

\newcommand{\method}{PrAda\xspace}

\newcommand{\myparagraph}[1]{\vspace{0.2cm}\noindent\textbf{#1}.\ }

\expandafter\def\expandafter\normalsize\expandafter{%
  \normalsize%
  \setlength\abovedisplayskip{2pt}%
  \setlength\belowdisplayskip{2pt}%
  \setlength\abovedisplayshortskip{2pt}%
  \setlength\belowdisplayshortskip{2pt}%
}

%% file: sec/0_abstract.tex
\begin{abstract}
  Segmenting images is critical for visual understanding but demands extensive pixel-level annotations. Foundational models have enabled new paradigms for predicting new classes guided by textual prompts, without annotations from the target domain. Yet, on specialized target domains, far from the original pre-training, their performance degrades.
  We study the errors of existing methods under such domain-shift, finding that misclassification rather than mask generation is the main culprit.
  To address this, we introduce the novel problem of Few-Shot Visual Adaptation for text-prompted Segmentation. This kind of adaptation has been largely studied for image classification, but it remains unexplored for segmentation.
  We tackle this task with \textbf{Pr}ototype \textbf{Ada}ptation (\method), a novel, parameter-efficient method that adapts a frozen text-prompted segmentation model. Our approach learns class-specific prototypes by combining fine-grained pixel features and high-level transformer representations, which are then fused with the original text-based predictions through a learned importance factor. This preserves the model’s zero-shot potential while enabling strong adaptation to new domains.
  Experiments across semantic, instance, and panoptic segmentation on five benchmarks demonstrate that \method yields significant improvements over state-of-the-art and proposed baselines. Code is available at {\small{\url{https://github.com/FocoosAI/PrAda}}}.
  \vspace{-1cm}
\end{abstract}

%% file: sec/1_intro.tex
\section{Introduction}
\label{sec:intro}

Segmentation is a fundamental computer vision task that focuses on dividing images into meaningful regions to enable a comprehensive understanding of scenes.
Unified segmentation models ~\cite{cheng2022masked, cheng2021per, li2023mask, cavagnero2024pem, kerssies2025your, rosi2024revenge}, which concurrently address all main paradigms (semantic, instance, and panoptic), have achieved outstanding performance across numerous applications. However, their broader adoption is constrained by the substantial cost of acquiring detailed pixel-level annotations \cite{bearman2016s}, particularly in specialized domains.

\begin{figure}[!t]
  \centering
  \includegraphics[width=1\linewidth]{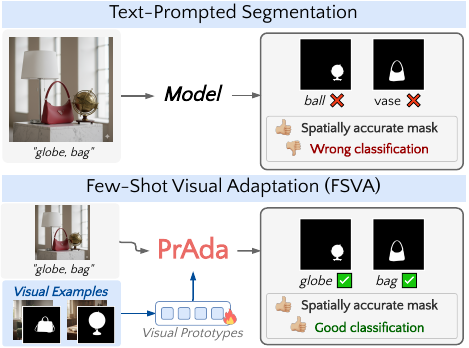}
  \caption{\textbf{Few-Shot Visual Adaptation for Text Prompted Segmentation (FSVA-Seg).} 
    Text-prompted segmentation models often localize object regions accurately, yet struggle to assign the correct category label. We introduce the FSVA-Seg setting, where just a few labeled examples are available to guide the adaptation process. Our approach, \method, learns class-specific visual prototypes from these examples to effectively boost classification performance while keeping the segmentation model frozen.
  }
  \label{fig:teaser}
\end{figure}

Recently, foundational models \cite{kirillov2023segment,radford2021learning, oquab2023dinov2} have emerged as a promising solution to mitigate the need for expensive pixel-level annotations. These models enable prompt-guided segmentation through diverse modalities: users can employ text prompts (\eg. natural language descriptions of the target objects) to enable open-vocabulary~\cite{zhang2023simple, xu2023open, yu2023convolutions} and referring segmentation~\cite{liu2023gres, cuttano2023cross}, or provide visual prompts (\eg. annotated reference images) to direct the model's predictions \cite{wang2023seggpt, wang2023images, li2024visual, cuttano2025samwise}.
Notably, segmentation models that use text prompts consistently achieve better performance than approaches that rely exclusively on visual prompts (as shown in \cref{sec:results}), highlighting the value of the high-level semantic information encoded in language.

However, current text-prompted segmentation models are fundamentally limited by their exclusive reliance on natural language directives, which often leads to sub-optimal performance in specialized domains or where precise verbal descriptions are unavailable \cite{rosi2025show}.
In the adjacent task of image classification, this issue has been successfully mitigated by few-shot visual adaptation, where text models are adapted using only a handful of annotated images \cite{zhou2022learning, gao2024clip, zhang2022tip, farina2025rethinking}. Crucially, this direction remains largely unexplored in the context of segmentation: a few works \cite{guiknn, guan2025text} explored the integration of text and visual prompts, but neither of them operates using only a small number of adaptation images.
In this work, we address this gap by introducing the \textbf{F}ew-\textbf{S}hot \textbf{V}isual \textbf{A}daptation  setting to text-prompted image \textbf{Seg}mentation (FSVA-Seg), with the goal of efficiently adapting text-prompted segmentation models using only a handful of examples from the target domain.

To inform the design of effective FSVA-Seg methods, we first set out to understand the failure mode of existing text-prompted models in specialized domains, asking the following question: \textit{do the primary errors stem from localization or classification?}
We conduct a comprehensive investigation taking as test-subject FC-CLIP \cite{yu2023convolutions}, a recent state-of-the-art model based on the Mask2Former \cite{cheng2022masked} paradigm, that is trained on a closed set of categories (COCO-Panoptic~\cite{lin2014microsoft}) and extended to out-of-vocabulary classes by leveraging a frozen CLIP encoder \cite{radford2021learning}. In particular, we evaluate qualitative and quantitative metrics on 28 domain-specific datasets and find that the model mostly localizes correctly the relevant regions. Instead, most errors arise from misclassifications, especially in domains semantically distant from the pretraining data (\cf. \cref{fig:teaser}).

Building on this insight, we introduce \textbf{Pr}ototype \textbf{Ada}ptation (\method), a novel parameter-efficient adaptation for a frozen FC-CLIP \cite{yu2023convolutions}. \method addresses the misclassification problem by learning class-specific prototypes, initialized from the aggregation of two complementary features sources: (i) fine-grained spatial features from the pixel decoder and (ii) high-level semantic representations derived from the transformer decoder query embeddings. During inference, we compute visual similarity scores between the prototypes and the image features. These scores are then fused with the original text-based classification score via a learned importance parameter that adaptively balances the two modalities. Only the class prototypes and the importance parameter are optimized using the few-shot examples, resulting in a lightweight adaptation. Furthermore, keeping the original model frozen preserves the ability to rely on CLIP for domains where textual prompts are effective.

We validate our method on semantic, instance, and panoptic segmentation across multiple benchmarks including ADE20K \cite{zhou2017scene}, Cityscapes \cite{cordts2016cityscapes}, Mapillary Vistas \cite{neuhold2017mapillary}, Segmentation-In-The-Wild \cite{zou2023generalized} and ShowOrTell \cite{rosi2025show}. For completeness, we evaluate the performance of \method also against 4 image-classification baselines \cite{zhou2022learning, zhou2022conditional, gao2024clip, zhang2022tip} that we extended for the FC-CLIP architecture.
Our approach demonstrates substantial improvements over the all the baselines, obtaining more than 4~PQ over FC-CLIP with only an increase in parameter cardinality of +0.02\% on Cityscapes \cite{cordts2016cityscapes} and +0.19\% on ADE20k \cite{zhou2017scene}.

In summary, the contributions of the paper are:
\begin{itemize}
  \item We propose the task of few-shot visual adaptation for text-prompted image segmentation (FSVA-Seg) and we reveal that the failure of text-prompted models on specialized domains is mostly linked to misclassifications.
  \item We introduce a novel lightweight approach, \method, that adapts a frozen FC-CLIP model through class-specific learnable prototypes that are optimized on a few visual examples.
  \item We evaluate our method against several proposed baselines on 5 benchmarks (28 datasets) and 3 tasks, demonstrating consistent improvements in performance.
\end{itemize}

%% file: sec/2_related.tex
\section{Related Work}
\label{sec:related}

\myparagraph{Segmentation with textual prompts}
Text-prompted segmentation leverages textual descriptions to guide the segmentation process, enabling flexible and semantic-aware object localization. Vision-Language Models (VLMs) like CLIP~\cite{radford2021learning} and ALIGN~\cite{jia2021scaling} are fundamental to this paradigm, thanks to their extensive vocabulary and visual-text alignment capabilities. In this field, two notable trends have emerged.
\textit{Two-stage methods}~\cite{ghiasi2022scaling,ding2022decoupling,xu2022simple,liang2023open,xu2023open}, which first generate class-agnostic mask proposals and then match them with CLIP-based text embeddings. This includes also methods that extend the text-guided formulation to panoptic and instance segmentation~\cite{jiao2024collaborative,xu2023open,yu2023convolutions}.
\textit{Single-stage frameworks}~\cite{li2022language,ding2022open,zhou2023zegclip,xu2023side,cho2024cat} bypass region proposals by predicting segmentation maps directly from CLIP embeddings, via learnable tokens or adapters.
Recent work~\cite{wysoczanska2024clip,wysoczanska2024clipdino,lan2024proxyclip,jose2025dinov2} combines VLMs with self-supervised vision models~\cite{caron2021emerging,oquab2023dinov2,simeoni2025dinov3} for improved localization.

\noindent
Differently, in this work we start by investigating the root cause of these models' failures when applied to specialized domains. Empirical observations highlighted the need for a stronger visual guidance which led us to design/formulate the first approach including visual adaptation from a few annotated images into a state-of-the-art text prompted segmentation model.

\myparagraph{Segmentation with visual prompts}
Visual prompting leverages geometric cues (\eg., points, bounding boxes, or masks) to guide localization of specific objects. This paradigm originates from object detection, where methods like OV-DETR~\cite{zang2022open} and OWL-ViT~\cite{minderer2022simple} employ CLIP encoders to process visual examples alongside text prompts for open-vocabulary detection. MQ-Det~\cite{xu2023multi} enriches textual descriptions using image exemplars, while T-Rex2~\cite{jiang2024t} unifies geometric and text prompts via region-level contrastive alignment, albeit treating visual prompts primarily as geometric auxiliary signals.
SAM~\cite{kirillov2023segment,ravi2024sam} establishes a milestone in interactive segmentation with decoupled visual prompt encoding, though it lacks semantic awareness. To mitigate this limitation, some recent works~\cite{zou2023segment,li2024visual} extend the framework by incorporating semantic context while maintaining the benefits of visual prompting.
However, existing segmentation methods mostly rely on visual clues, without synergizing them with the textual modality.

\noindent
Unlike prior work, we enhance segmentation performance (semantic, instance and panoptic) by leveraging the guidance of both textual prompts and visual examples.

\myparagraph{Few-shot Visual Adaptation}
The problem of adapting VLMs with only a few annotated visual examples has been explored in image classification~\cite{xing2024survey,wortsman2022robust}.
While traditional fine-tuning~\cite{brown2020language,devlin2019bert,he2016deep} updates all parameters, computational burden and the risk of overfitting have motivated PEFT strategies~\cite{houlsby2019parameter,hu2022lora,li2021prefix}, extended to VLMs~\cite{gao2024clip,khattak2023maple,zhou2022learning,zhou2022conditional}.
\textit{Prompt learning} provides textual instructions to enhance task comprehension.
CoOp~\cite{zhou2022learning} optimizes learnable context vectors in the language branch, while CoCoOp~\cite{zhou2022conditional} generates instance-conditioned prompts via a lightweight network.
Other works preserve knowledge through gradient projection~\cite{zhu2023prompt} or construct multi-modal prompts~\cite{khattak2023maple,chen2022plot}.
\textit{Adapter-based methods} offer an alternative strategy.
CLIP-Adapter~\cite{gao2024clip} combines pretrained and adapted features through non-linear transformations, while Tip-Adapter~\cite{zhang2022tip} uses a cache-based approach without fine-tuning.
CLIP-LoRA~\cite{hu2022lora} introduces low-rank adaptation matrices~\cite{yu2023task,yang2024mma}.
\textit{Two-stage methods}~\cite{farina2025rethinking} first fine-tune a feature extractor, then train a classifier on top.

\noindent
While effective for classification, only few works explored how to exploit both textual and visual prompts in segmentation: kNN-CLIP \cite{guiknn} uses a retrieval approach upon a frozen FC-CLIP, while Prompt-DINO \cite{guan2025text} builds a fusion mechanism upon DETR \cite{carion2020end}. However, both require sizable amount of annotated data, leaving the field of few-shot adaptation in segmentation unexplored.

%% file: sec/3_method.tex
\section{Few-shot visual adaptation for text-prompted image segmentation}
\label{sec:method}

\myparagraph{Task Definition}
We consider the general task of segmentation (semantic, instance or panoptic): given an input image $I \in \mathbb{R}^{H \times W \times 3}$, we aim to predict all segments in $I$ from a finite set $\mathcal{C}$ of categories.
Formally, we seek a map $f_\theta$, with learnable parameters $\theta$, such that
\begin{equation}
  f_\theta(I) \mapsto \{(\hat{m}_i, \hat{c}_i)\}_{i=1}^N
\end{equation}
where $\hat{m}_i \in \{0, 1\}^{H \times W}$ is a binary mask and $\hat{c}_i\in\mathcal{C}$ is its corresponding category.

In text-prompted segmentation \cite{li2022language,ding2022open,zhou2023zegclip,xu2023side,cho2024cat} the set of categories $\mathcal{C}$ determines the textual prompts $\mathcal{T} = \{t_i\}_{i=1}^{T}$ that guide the segmentation. In this work, we introduce a new setting where we assume to additionally have a small number of annotated images from a target domain, as reference. These visual annotations are given as binary masks with an associated class label.
Formally, we denote the set of all these visual examples as $\mathcal{V}=\{v_i\}_{i=1}^{V}$, with $v_i=(m_i,c_i)$ being a pair composed of a binary mask $m_i \in \{0, 1\}^{H \times W}$ and a class label $c_i \in \mathcal{C}$ (\cf. \cref{fig:prototypes_extraction}). We omit in this notation the reference images to which the annotations correspond to, for the sake of readability.
Note that this formulation generalizes to reference images containing a single or multiple visual examples.

\myparagraph{Background}
Our approach (\cf. \cref{fig:method}) builds upon FC-CLIP \cite{yu2023convolutions}, an efficient framework for open-vocabulary segmentation
that pairs a frozen CLIP \cite{radford2021learning} vision encoder with Mask2Former's \cite{cheng2022masked} pixel and mask decoders, trained on COCO-Panoptic~\cite{lin2014microsoft}.
In particular, a set of $N$ learnable queries $Q \in \mathbb{R}^{N \times D}$ is first gradually refined by passing through a set of cross-attention layers together with the visual features extracted from the pixel-decoder. This enriches the queries with semantic and spatial information.
Then, the $i$-th mask $\hat{m}_i$ is inferred as
\begin{equation}
  \label{eq:mask_prediction}
  \hat{m}_i = F \cdot q_i
\end{equation}
where $q_i \in Q$ is the $D$-dimensional query and $F \in \mathbb{R}^{H \times W \times D}$ indicates the visual features\footnote{The visual features are taken from the last layer of the pixel-decoder, after an upsampling.}.
Finally, the corresponding category $\hat{c}_i$ is predicted by the similarity between the textual embeddings generated from the prompts $\mathcal{T}$ using CLIP's text encoder and a set of $N$ class embeddings.

\subsection{What limits the generalization capabilities?}
\label{sec:classification_problem}
Despite leveraging a frozen CLIP encoder, which has emergent zero-shot capabilities \cite{radford2021learning}, FC-CLIP's performance deteriorates when generalizing to categories that are not in the training set (COCO-Panoptic).
This suboptimal behavior on novel categories can be attributed to: (1) the feature representations learned during training may not be sufficiently discriminative for novel categories, leading to poor mask localization; (2) the classification mechanism itself may not be robust enough to accurately associate textual prompts with visual features for novel categories, leading to poor classification accuracy.

To gauge the impact of these two factors, we conduct a thorough quantitative and qualitative analysis of FC-CLIP's predictions on 28 datasets.
First, we inspect all the $N$ predicted binary masks and perform an IoU-based matching with the ground truth annotations of the datasets (\cf. \cref{fig:motivations}-a).  We find that the model is able to generate spatially accurate masks, but it frequently assigns incorrect category labels.
Additionally, we observe that a significant portion of the predicted masks achieves IoU scores around 0.8 with ground truth annotations. This indicates that the mask decoder is capable of robust, class-agnostic object localization based on visual cues, while the main source of error lies in the classification.
To confirm this hypothesis, we further perform an oracle experiment replacing the predicted classification scores with ground truth classes. The results (\cf. \cref{fig:motivations}-b) show a significant boost across various datasets when using ground truth labels. Results on all 28 datasets are reported in the supplementary material.

Our comprehensive analysis indicates the mask classification as the primary bottleneck. Unlike prior work that presented results only on narrow gaps (\eg, target domain ADE20K \cite{zhou2017scene}), our experiments conclusively show that these findings are consistent on more challenging distributions, such as the real-world benchmark SegInW \cite{zou2023generalized}.

\subsection{Few-shot adaptation using prototypes}
\label{sec:prototypes}
Having identified the main cause of FC-CLIP's degraded performance on the unknown semantics of novel categories, we set out to improve the mask classification by leveraging the few visual examples $\mathcal{V}$ collected from the target domain.
A naive mitigation strategy would be to directly fine-tune the model on the given samples. However, this leads to catastrophic forgetting \cite{li2017learning, cermelli2023comformer}, worsening the model's performance on the original training categories and hindering its generalization capabilities. There is also a well established literature of few-shot adaptation strategies in image classification, which addresses the same problem by adapting the model's visual representation \cite{gao2024clip,zhang2022tip,khattak2023maple,yang2024mma}.
However, extending these methods to segmentation is not straightforward. While they typically operate on the global representations produced by CLIP's text and visual encoders, segmentation models are inherently more complex, as they rely on pixel- or segment-level features rather than a single feature for the entire image.

Hence, we propose a novel approach tailored for image segmentation, called \method. The underlying idea is to
integrate the textual similarity score of a frozen FC-CLIP with a visual similarity \wrt a set of prototypes derived from the examples $\mathcal{V}$. An overview of \method is reported in \cref{fig:method}.

In the rest of this section, we first describe in detail our process to initialize these prototypes from the frozen FC-CLIP feature space.
Then, we discuss how these prototypes are fine-tuned on the provided visual examples, which is crucial to capture the nuances of the novel categories, considering that the target may differ significantly from the original training distribution.
Lastly, we explain the learnable balancing between text and visual prompts, in order to produce the final predictions.

\begin{figure}[t]
  \centering
  \includegraphics[width=1\linewidth]{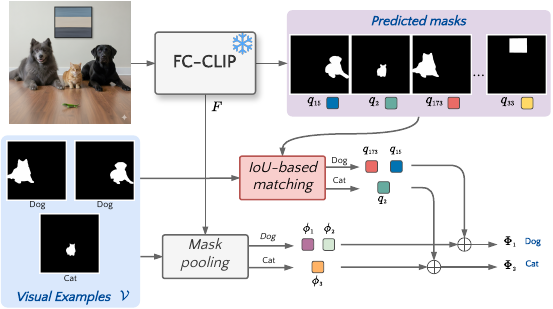}
  \caption{\textbf{Prototypes initialization}. Given and image first we employ the frozen version of FC-CLIP \cite{yu2023convolutions} to predict a series of mask where each is associated with a query $q_{i}$. Given the set of visual example we perform (1) an IoU-based matching to obtain the queries relative to the predicted mask that better match with the visual examples and (2) a mask pooling operation to obtain a condensed feature representation. The final class prototypes are given by the sum of these two vectors.}
  \label{fig:prototypes_extraction}
\end{figure}

\myparagraph{Prototypes initialization}
The key to our approach is obtaining prototypes that summarize well the visual and semantic characteristics of the sought categories $\mathcal{C}$, by leveraging the visual examples $\mathcal{V}$.
This entails understanding what is the most suitable visual representation attainable from the model.
We start by considering the pixel-level features $F \in \mathbb{R}^{H \times W \times D}$ obtained from the pixel-decoder. These dense features, which are used by the model to correctly predict class-agnostic masks via \cref{eq:mask_prediction}, capture fine-grained visual details which are also important for recognizing categories.
For each visual example in $v_i \in \mathcal{V}$, we condense these features into a vector representation $\phi_i$ using the mask average pooling operation \cite{xu2023open}:
\begin{equation}
  \label{eq:visual_representation}
  \phi_i = \frac{\sum_{j=1}^{H\cdot W} m_{i,j} F_{j}}{\sum_{j=1}^{H \cdot W} m_{i,j}},
\end{equation}
where the subscript ${j}$ notes the pixel location in the image.

Although the pooled feature $\phi_i$ summarizes the appearance of the category represented in the visual example $v_i$, it is not specialized to capture semantic information.
This specialization is instead present in the queries $Q$, which in the transformer decoder are jointly optimized for mask prediction and classification, thus learning to capture high-level semantic information about the category they represent. Therefore, we propose to combine the visual information in $\phi_i$ with the semantic information in the query embeddings. However, we lack a correspondence between individual queries and the reference masks. To establish this correspondence, we first pass the reference image through the frozen FC-CLIP to compute a set of $N$ predicted masks (\cf. \cref{fig:prototypes_extraction}). Then, we perform  an IoU-based matching between the reference mask $m_i$ and all the predicted masks. Finally, we assign to the visual example $v_i$ the query that generated the best-matching prediction. We indicate this query as $q_i$, with a little abuse of notation\footnote{Without loss of generality, we assume a permutation such that the $i$-th query corresponds to the $i$-th visual example.}.

Having computed the pooled visual features and corresponding query for each visual example $v_i$, we aggregate them to generate a single prototype for each category in $\mathcal{C}$. Denoting with $\mathcal{V}_k \subset \mathcal{V}$ the subset of visual examples pertaining the $k$-th category in $\mathcal{C}$, its prototype $\Phi_k \in \mathbb{R}^D$ is
\begin{equation}
  \Phi_k = \frac{1}{|\mathcal{V}_k|} \sum_{i : v_i\in \mathcal{V}_k}
  \left( q_i + \phi_i\right), \quad k=1,..,|\mathcal{C}|.
\end{equation}
For compactness, we stack all the prototypes in a single matrix $\Phi \in \mathbb{R}^{|C| \times D}$.

\myparagraph{Prototype-based visual similarity}
The set of prototypes $\Phi$ guides the classification by summarizing the appearance and high-level semantics of the target categories $\mathcal{C}$.
To use this guidance, at inference we forward the target image $I$ through the model and get its set of predicted masks $\{\hat{m}_i\}_{i=1}^N$. Each predicted mask is, by construction, associated to a query $q_i$. For each of these predicted masks, we compute a visual representation $\hat{\phi}_i$ using \cref{eq:visual_representation} with $\hat{m}_i$ instead of the reference mask.
Finally, we compute the visual representation for each predicted mask as
\begin{equation}
  \hat{\Phi}_i = q_i + \hat{\phi}_i
\end{equation}
and stack them all into a matrix $\hat{\Phi}\in\mathbb{R}^{N \times D}$.

With this, we can ultimately compute a similarity score between the visual representations of the predicted masks and the prototypes obtained from the examples $\mathcal{V}$ by simply taking their cosine similarity
\begin{equation}
  \label{eq:svisual}
  S_{\text{visual}} = \frac{\hat{\Phi} \cdot \Phi^T}{\|\hat{\Phi}\|_2 \|\Phi\|_2}.
\end{equation}
In practice, to account for the void category during classification, we augment the set of prototypes with an additional prototype that captures the ``no-class'' category. Therefore, $S_{\text{visual}}$ is actually a $N \times (|\mathcal{C}|+1)$ dimensional matrix.

\myparagraph{Fine-tuning the prototypes}
The initialization of the prototypes $\Phi$ summarizes the visual and semantic information of examples $\mathcal{V}$ according to the frozen feature space of the base model.
However, since the decoder was trained on a different data distribution, this feature space may not optimally capture the discriminative patterns and nuances specific to the target categories. 
Hence, we fine-tune the prototypes on the
provided visual examples while keeping the FC-CLIP model frozen.
This strategy allows the prototypes to adapt beyond the constraints of the frozen feature space without risking catastrophic forgetting of the base model's knowledge.

During fine-tuning, we compute the visual similarity scores $S_{\text{visual}}$ as in \cref{eq:svisual} and combine them with the textual similarity $S_{\text{text}}$ from the frozen FC-CLIP.
However, finding the optimal balance between textual prompts $\mathcal{T}$ and visual examples $\mathcal{V}$ is non-trivial, as their relative effectiveness varies across domains \cite{rosi2025show,jiang2024t}.
Thus, we introduce a learnable scalar $\alpha$ to adaptively weigh the two signals:
\begin{equation}
  \label{eq:final_score}
  S_{\text{final}} = S_{\text{text}} + \alpha S_{\text{visual}}.
\end{equation}
We optimize only $\Phi$ and $\alpha$ using a cross-entropy loss on $S_{\text{final}}$, following \cite{yu2023convolutions}.
The combined score in \cref{eq:final_score} is also used at inference, to predict the class of the masks.

%% file: sec/4_results.tex
\section{Experiments}
\label{sec:results}

\input{tables/main_results.tex}

\myparagraph{Datasets and Metrics} We begin with a frozen FC-CLIP~\cite{yu2023convolutions} model pre-trained on COCO-Panoptic~\cite{lin2014microsoft} and adapt it to a broad range of datasets. Specifically, we use:
\textbf{ADE20K}~\cite{zhou2017scene}, the standard benchmark for segmentation, comprising 150 classes collected in a wide variety of everyday scenes;
\textbf{Cityscapes}~\cite{cordts2016cityscapes}, which focuses on urban driving environments with 19 classes;
\textbf{Mapillary Vistas}~\cite{neuhold2017mapillary}, a benchmark capturing 66 classes across geographically and visually diverse road scenes.
Additionally, for a better assessment of real-world generalization, we report results on \textbf{SegInW}~\cite{zou2023generalized}, a comprehensive suite of 25 instance segmentation datasets, and \textbf{ShowOrTell} (\textbf{SoT})~\cite{rosi2025show}, which aggregates 14 semantic segmentation datasets from 7 distinct domains. Together, these benchmarks allow us to evaluate adaptation efficacy across a challenging spectrum of targets.
In line with previous works~\cite{yu2023convolutions, li2024visual}, we evaluate Panoptic Quality (PQ)~\cite{kirillov2019panoptic} for panoptic segmentation, mean Intersection over Union (mIoU)~\cite{everingham2015pascal} for semantic segmentation, and Average Precision (AP)~\cite{lin2014microsoft} for instance segmentation. 

\myparagraph{Baselines}
Given the novelty of the FSVA-Seg setting, we implement a series of baselines by carefully re-purposing leading CLIP-based classification techniques for the FC-CLIP~\cite{yu2023convolutions} architecture. Specifically, we consider: (1) \textbf{CoOp} and (2) \textbf{CoCoOp}, which replace fixed text prompts with learnable ones in accordance with their original designs; (3) \textbf{CLIP-Adapter}~\cite{gao2024clip}, which is integrated by appending the adapter after the computation of class embeddings (rather than the [CLS] token); (4) \textbf{TIP-Adapter}~\cite{zhang2022tip}, which we include in its fine-tuned variant (\textit{TipAdapter-F}). For TIP-Adapter, the cache comprises the mask-pooled pixel features generated by the segmentation head, while all subsequent scoring steps mirror the original approach.

\myparagraph{Implementation Details}
All methods are implemented with the FC-CLIP~\cite{yu2023convolutions} framework, using two CLIP~\cite{radford2021learning} backbone variants: ResNet-50~\cite{he2016deep} and ConvNeXt-L~\cite{liu2022convnet}. We denote our method as \method-R50 when using CLIP-R50 and \method-L when using ConvNeXt-L. For adaptation, we randomly select 5 images per class from each dataset's training split. To ensure reproducibility and robustness to sampling, we report average results over 5 random seeds. All training and adaptation protocols, including hyperparameter choices and data processing, are held fixed across all methods for fair comparison, with additional details provided in the supplementary material. For evaluation, we use the same protocol of FC-CLIP \cite{yu2023convolutions} on the datasets validation splits. Code will be made publicly available upon acceptance.

\input{tables/challenging_results.tex}

\subsection{Results}
\myparagraph{Results on ADE20K} We begin by comparing our method (\method-L) to state-of-the-art baselines on ADE20K, which has relatively little domain shift from COCO. As shown in \cref{tab:dinov-benchmark}, \method-L outperforms most baselines: in particular, it improves over FC-CLIP by 4.6~PQ and 4.1~mIoU, and exceeds DINOv-L by even larger margins while using fewer adaptation samples (+8.2~PQ). \method-L yields the best PQ among all methods except Prompt-DINO, which uses stronger pre-training and does not disclose the number of visual prompts used during inference. In addition, it outperforms all the methods on mIoU except kNN-CLIP, that obtains 2.9~mIoU more than \method-L but uses the \textit{full} target dataset.
Compared to few-shot visual adaptation baselines, our approach consistently surpasses all competitors on panoptic and semantic metrics: it exceeds CLIP-Adapter by 0.3~PQ and 0.7~mIoU, TipAdapter-F by 1.1~PQ, and CoOp by 4.9~PQ and 6.6~mIoU. For instance segmentation, \method-L achieves 18.1~AP, which exceeds CoOp (16.4~AP) but falls slightly behind CLIP-Adapter (19.2~AP). This may be attributed to a better calibration of CLIP-Adapter, since it achieves higher AP but falls behind in other metrics.
Considering the ResNet-50 versions, \method outperforms all the baselines by higher margin, suggesting a better effectiveness with smaller model capacity.

\myparagraph{Results on Street View Datasets} On Cityscapes and Mapillary, which exhibit substantial diversity with respect to the pre-training dataset, our method demonstrates more pronounced advantages.
Among the SotA methods, FC-CLIP obtains the best performance on both datasets, even \wrt methods pre-trained on much larger datasets (+9.7~PQ, 4.0~mIoU \wrt Prompt-DINO and +11.4~PQ, 19.1~mIoU \wrt APE-L on Cityscapes). This demonstrates that properly using the language prompts is essential to generalize well to unseen domains. Nevertheless, \method improves on FC-CLIP by +5.8~PQ and +10.0~mIoU on Cityscapes, and +5.4~PQ and +10.2~mIoU on Mapillary, confirming the effectiveness of the adaptation.
Compared to FSVA-Seg baselines, our approach achieves consistent improvements on both Cityscapes and Mapillary. Specifically, \method-L (\method-R50) outperforms CLIP-Adapter by 1.2~PQ (0.7~PQ) on Cityscapes and by 3.8~PQ (4.1~PQ) on Mapillary, while also yielding gains over TipAdapter-F and CoCoOp on both datasets. We observe similar trends in semantic segmentation performance, with our method surpassing the best baseline by up to 1.9~mIoU (0.6~mIoU) on Cityscapes and 9.4~mIoU (7.9~mIoU) on Mapillary. These results demonstrate that tailoring few-shot adaptation specifically for segmentation delivers superior performance in domains that are distant from the original pre-training.

\myparagraph{Generalization to Challenging Scenarios}
As shown in \cref{tab:seginw-results}, on SegInW \method-L achieves 43.6~AP, outperforming the zero-shot FC-CLIP-L baseline (41.6~AP) and the visual reference method DINOv-L (40.6~AP) which requires 16 images per class. Our method also surpasses strong text-based models like X-Decoder-L when fine-tuned on 5 examples per class of the target datasets (35.5~AP). Even though APE-L and Prompt-DINO-L exceeds the performance of \method-L, we note that they have been pretrained on a larger dataset corpus, obtaining better generalization performance. In addition, as discussed previously, Prompt-DINO-L do not disclose the adaptation protocol and number of visual prompts, making difficult a proper comparison.
\method-L outperforms the FSVA-Seg baselines by a large margin, obtaining 1.2~AP more than CLIP-Adapter.

On the SoT benchmark \method-L achieves 33.1 mIoU, as shown in Table~\ref{tab:sot-results}. This result is comparable with other SotA methods such as Matcher (33.5 mIoU), and only worse than GFSAM (38.7 mIoU) which however require much more computation. In particular, it uses two foundational models, DINOv2 \cite{oquab2023dinov2} and SAM \cite{kirillov2023segment}, and predicts the output one class at at time, as detailed in \cite{rosi2025show}. On the contrary, \method-L requires a single forward with a simpler architecture.
\method-L surpasses FC-CLIP by 10.0~mIoU, clearly indicating the benefits of few-shot visual adaptation. Furthermore, when comparing with other FSVA-Seg baselines, it obtains results better by a wide margin: +5.4 and +6~mIoU \wrt CLIP-Adapter and Tip-Adapter-F, respectively.

\subsection{Ablation Studies}

\myparagraph{Prototypes representation} We ablate different methods to build prototypes based on the ConvNext in \cref{tab:ablation-prototypes}. Even using random prototypes gives a notable boost over not using prototypes at all (26.8~PQ vs 14.6~PQ on ADE20K, and 48.4~PQ vs 44.2~PQ on Cityscapes). When using class embeddings, we see further improvements (30.1~PQ ADE20K, 48.4~PQ Cityscapes), but at the cost of significantly more parameters (116K~vs~39K). Applying mask pooling or queries alone raises performance, with queries are slightly better than mask pooling (31.7~PQ vs 28.9~PQ on ADE20K; equal 49.3~PQ on Cityscapes). Our approach, which combines queries and mask pooling, achieves the strongest results (32.2~PQ ADE20K, 50.1~PQ Cityscapes) while keeping the parameter footprint low, demonstrating the complementary benefits of object-centric and spatial information in learning effective prototypes for adaptation.

\input{tables/ablation_prototypes.tex}

\begin{figure}[t]
  \centering
  \includegraphics[width=0.9\linewidth]{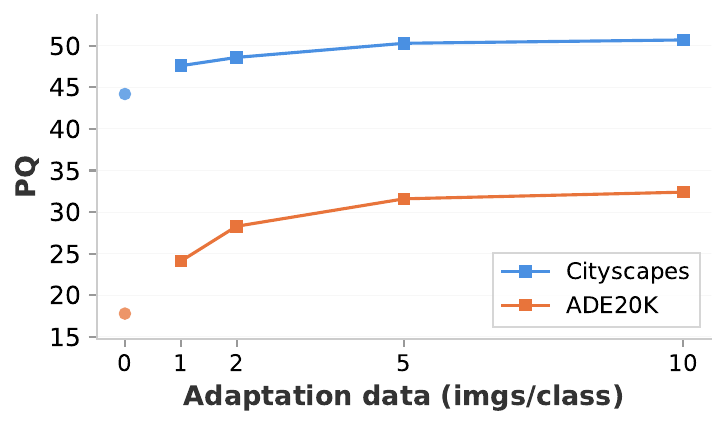}
  \caption{\textbf{Influence of the number of adaptation images per class.} We evaluate our approach on ADE20K and Cityscapes using 2, 5, and 10 adaptation images per class, and compare it to the zero-shot baseline without prototypes.}
  \label{fig:ablation_num_images}
\end{figure}

\myparagraph{Impact of adaptation data} We ablate the influence of the number of adaptation images per class by reporting PQ on both Cityscapes~\cite{cordts2016cityscapes} and ADE20K~\cite{zhou2017scene} across five configurations: 0 (\ie no prototypes), 1, 2, 5, and 10 images per class (\cref{fig:ablation_num_images}). On Cityscapes, our method improves steadily from 44.2 PQ in the zero-shot setting, to 47.2 with just 1 image, 48.1 with 2 images, 49.8 with 5 images, and 50.7 with 10 images per class. On ADE20K, we observe a similar trend, with results rising from 14.6 (zero-shot), to 24.5 (1 image), 28.7 (2 images), 31.4 (5 images), and reaching 33.2 (10 images). These results highlight the strong sample efficiency of our method: even with a single adaptation sample per class, we obtain sizable gains over the baseline, and steady improvements are observed as more examples are added, though with diminishing returns beyond 5 images per class.

%% file: tables/main_results.tex
\begin{table*}[t]
  \setlength{\tabcolsep}{3pt}
  \renewcommand{\arraystretch}{1.0}
  \centering
  \resizebox{\textwidth}{!}{
    \begin{tabular}{l c c c c c c c c c c}
      \toprule
      \multirow{2}{*}{Model} & \multirow{2}{*}{
        \begin{tabular}{c}Prompt\\Type
      \end{tabular}} & \multirow{2}{*}{Pre-Training Data} & \multirow{2}{*}{Adaptation Data} & \multicolumn{3}{c}{ADE20K} & \multicolumn{2}{c}{Mapillary} & \multicolumn{2}{c}{Cityscapes} \\
      \cmidrule(lr){5-7} \cmidrule(lr){8-9} \cmidrule(lr){10-11}
      & & & & PQ & mIoU & AP & PQ & mIoU & PQ & mIoU \\

      \midrule
      FC-CLIP-R50 \cite{yu2023convolutions}        & T & COCO-Panoptic & - & 17.9 & 23.3 & 9.5 & 15.9 & 24.4 & 40.3 & 53.2 \\
      DINOv-T \cite{li2024visual}            & V & COCO, SA-1B & 16 imgs/class & 19.4 & 21.9 & 11.4 & - & - & - & -  \\
      OpenSEED-L \cite{zhang2023simple}         & T & COCO + O365 & - & 19.7 & 23.4 & 15.0 & - & - & 41.4 & 47.8\\
      X-Decoder-L \cite{zou2023generalized}         & T & COCO + CC3M + \dots & - & 21.8 & 29.6 & 13.1 & - & - & 38.1 & 52.0 \\
      ODISE \cite{xu2023open}       & T & COCO-Panoptic & - & 22.6 & 29.9 & 14.4 & 14.2 & - & 23.9 & - \\
      DINOv-L \cite{li2024visual}            & V & COCO, SA-1B & 16 imgs/class & 23.2 & 25.3 & 15.1 & -  & - & - & - \\
      APE-L \cite{shen2024aligning}              & T & \scriptsize{COCO + O365 + OpenImages + \dots} & - & 26.6 & 28.9 & 23.8 & - & - & 28.6 & 37.9 \\
      FC-CLIP-L \cite{yu2023convolutions}    & T & COCO-Panoptic & - & 26.8 & 34.1 & 16.8 & 18.2 & 27.9 & 44.0 & 56.2 \\
      MAFT-L \cite{jiao2024collaborative}    & T & COCO-Panoptic  & - & 27.1 & 36.1 & -  & - & - & - & -  \\
      kNN-CLIP \cite{guiknn}                              & T+V & COCO-Panoptic & \textit{Full dataset} & 29.6 & 41.3 & 17.5 & - & - & - & - \\
      Prompt-DINO-L \cite{guan2025text} & T & \scriptsize{COCO + O365 + CC3M + SA-1B + \dots} & - & 29.6 & - & 24.6 & - & - & - & - \\
      Prompt-DINO-L \cite{guan2025text} & V & \scriptsize{COCO + O365 + CC3M + SA-1B + \dots} & ND & 35.9 & - & 28.0 & - & - & 34.3 & 52.2 \\

      \midrule
      \multicolumn{11}{l}{\textbf{Few-Shot Visual Adaptation}} \\
      FC-CLIP-R50 + CoOp \cite{zhou2022learning}          & T+V & COCO-Panoptic & 5 imgs/class & 19.8{\footnotesize $\pm$0.3} & 23.9{\footnotesize $\pm$0.7} & 10.2{\footnotesize $\pm$0.3} & 15.7{\footnotesize $\pm$0.6} & 20.6{\footnotesize $\pm$1.2} & 41.5{\footnotesize $\pm$1.8} & 51.9{\footnotesize $\pm$2.8} \\

      FC-CLIP-R50 + CLIP-Adapter \cite{gao2024clip}      & T+V & COCO-Panoptic & 5 imgs/class & 21.6{\footnotesize $\pm$0.3} & 26.9{\footnotesize $\pm$0.4} & 11.6{\footnotesize $\pm$0.2} & 17.9{\footnotesize $\pm$0.1} & 26.6{\footnotesize $\pm$0.1} & 44.8{\footnotesize $\pm$0.7} & 60.8{\footnotesize $\pm$1.0} \\
      FC-CLIP-R50 + TipAdapter-F \cite{zhang2022tip}     & T+V & COCO-Panoptic & 5 imgs/class & 21.4{\footnotesize $\pm$0.2} & 28.1{\footnotesize $\pm$0.3} & 10.6{\footnotesize $\pm$0.1} & 18.3{\footnotesize $\pm$0.0} & 27.6{\footnotesize $\pm$0.0} & 44.7{\footnotesize $\pm$0.2} & 59.6{\footnotesize $\pm$0.5} \\
      \textbf{\method-R50 }                                        & T+V & COCO-Panoptic & 5 imgs/class & \textbf{26.0}{\footnotesize $\pm$0.5} & \textbf{31.6}{\footnotesize $\pm$0.5} & \textbf{14.5}{\footnotesize $\pm$0.2} & \textbf{22.0}{\footnotesize $\pm$0.2} & \textbf{34.7}{\footnotesize $\pm$0.3} & \textbf{45.5}{\footnotesize $\pm$0.8} & \textbf{61.4}{\footnotesize $\pm$1.4} \\

      \arrayrulecolor{black!30}\midrule
      \arrayrulecolor{black}
      FC-CLIP-L + CoOp \cite{zhou2022learning}            & T+V & COCO-Panoptic & 5 imgs/class & 26.5{\footnotesize $\pm$0.2} & 31.6{\footnotesize $\pm$1.4} & 16.4{\footnotesize $\pm$0.2} & 16.9{\footnotesize $\pm$0.9} & 20.6{\footnotesize $\pm$1.2} & 46.8{\footnotesize $\pm$0.9} & 59.3{\footnotesize $\pm$1.7} \\

      FC-CLIP-L + CoCoOp \cite{zhou2022conditional}       & T+V & COCO-Panoptic & 5 imgs/class & 26.4{\footnotesize $\pm$0.5} & 31.9{\footnotesize $\pm$1.1} & 16.1{\footnotesize $\pm$0.4} & 14.2{\footnotesize $\pm$0.3} & 19.3{\footnotesize $\pm$0.8} & 46.9{\footnotesize $\pm$0.3} & 60.5{\footnotesize $\pm$1.7} \\
      FC-CLIP-L + CLIP-Adapter \cite{gao2024clip}         & T+V & COCO-Panoptic & 5 imgs/class & 31.1{\footnotesize $\pm$0.1} & 37.5{\footnotesize $\pm$0.2} & \textbf{19.2}{\footnotesize $\pm$0.1} & 19.8{\footnotesize $\pm$0.0} & 28.7{\footnotesize $\pm$0.2} & 48.6{\footnotesize $\pm$0.5} & 64.3{\footnotesize $\pm$0.3} \\
      FC-CLIP-L + TipAdapter-F \cite{zhang2022tip}        & T+V & COCO-Panoptic & 5 imgs/class & 30.3{\footnotesize $\pm$0.1} & \textbf{38.5}{\footnotesize $\pm$0.3} & 17.9{\footnotesize $\pm$0.1} & 20.5{\footnotesize $\pm$0.0} & 31.0{\footnotesize $\pm$0.1} & 48.0{\footnotesize $\pm$0.2} & 63.2{\footnotesize $\pm$0.5} \\

      \textbf{\method-L }                                    & T+V & COCO-Panoptic & 5 imgs/class & \textbf{31.4}{\footnotesize $\pm$0.7} & \textbf{38.2}{\footnotesize $\pm$0.5} & 18.1{\footnotesize $\pm$0.4} & \textbf{23.6}{\footnotesize $\pm$0.1} & \textbf{38.1}{\footnotesize $\pm$0.3} & \textbf{49.8}{\footnotesize $\pm$0.5} & \textbf{66.2}{\footnotesize $\pm$1.5} \\

      \bottomrule
  \end{tabular}}
  \caption{\textbf{Evaluation on standard datasets.} Few-Shot Visual Adaptation results across ADE20K~\cite{zhou2017scene}, Cityscapes~\cite{cordts2016cityscapes} and Mapillary~\cite{neuhold2017mapillary} benchmarks. We carefully implemented the few-shot adaptation baselines while we report official results for the other approaches. Note that Prompt-DINO \cite{guan2025text} does not disclose how many images are used for visual prompting.}
  \label{tab:dinov-benchmark}
\end{table*}

%% file: tables/challenging_results.tex
\begin{table}[t]
  \setlength{\tabcolsep}{3pt}
  \renewcommand{\arraystretch}{1.0}
  \centering
  \resizebox{\columnwidth}{!}{
    \begin{tabular}{l c c c c}
      \toprule
      \multirow{2}{*}{Model} & \multirow{2}{*}{
        \begin{tabular}{c}Prompt\\Type
      \end{tabular}} & \multirow{2}{*}{
        \begin{tabular}{c}Pre-training\\Data
      \end{tabular}} & \multirow{2}{*}{
        \begin{tabular}{c}Adaptation\\Data
      \end{tabular}} & SegInW \cite{zou2023generalized} \\
      \cmidrule(lr){5-5}
      & & & & mAP \\

      \midrule
      OpenSEED-L  \cite{xu2023open}        & T & COCO,O365 & - & 15.2 \\
      X-Decoder-L  \cite{zou2023generalized}         & T & COCO,... & - & 32.3 \\

      DINOv-L \cite{li2024visual}            & V & COCO, SA-1B & 16 imgs/class & 40.6 \\
      FC-CLIP-L \cite{yu2023convolutions}  & T & COCO-Panoptic & - & 41.6 \\
      APE-L \cite{shen2024aligning}              & T & \scriptsize{COCO,LVIS,O365,OI,VG} & - & {47.4} \\
      Prompt-DINO-L \cite{guan2025text}  & ND & \scriptsize{COCO,LVIS,O365,...} & ND & 56.4 \\

      \midrule
      \multicolumn{5}{l}{\textbf{Few-Shot Visual Adaptation}} \\
      X-Decoder-L (FT) \cite{zou2023generalized}        & T+V & COCO,... & 5 imgs/class & 35.5 \\
      FC-CLIP-L + CoOp \cite{zhou2022learning}             & T+V & COCO-Panoptic & 5 imgs/class & 41.2{\footnotesize $\pm$0.8} \\
      FC-CLIP-L + CoCoOp \cite{zhou2022conditional}        & T+V & COCO-Panoptic & 5 imgs/class & 41.5{\footnotesize $\pm$0.5} \\
      FC-CLIP-L + CLIP-Adapter \cite{gao2024clip}          & T+V & COCO-Panoptic & 5 imgs/class & 42.1{\footnotesize $\pm$0.4} \\
      \textbf{\method-L}              & T+V & COCO-Panoptic & 5 imgs/class & 43.3{\footnotesize $\pm$0.6} \\

      \bottomrule
  \end{tabular}}
  \caption{\textbf{Evaluation on SegInW} \cite{zou2023generalized}. We carefully implemented the few-shot adaptation baselines while we report official results for the other approaches. X-Decoder-L (FT) denotes it has been fine-tuned on the target datasets.}
  \label{tab:seginw-results}
\end{table}

\begin{table}[t]
  \setlength{\tabcolsep}{3pt}
  \renewcommand{\arraystretch}{1.0}
  \centering
  \resizebox{\columnwidth}{!}{
    \begin{tabular}{l c c c c}
      \toprule
      \multirow{2}{*}{Model} & \multirow{2}{*}{
        \begin{tabular}{c}Prompt\\Type
      \end{tabular}} & \multirow{2}{*}{
        \begin{tabular}{c}Visual\\Encoder
      \end{tabular}} & \multirow{2}{*}{
        \begin{tabular}{c}Adaptation\\Data
      \end{tabular}} & SoT \cite{rosi2025show} \\
      \cmidrule(lr){5-5}
      & & & & mIoU \\

      \midrule
      SINE \cite{liu2024simple}   & V & DINOv2~\cite{oquab2023dinov2} & 5 imgs/class & 23.2 \\
      FC-CLIP \cite{yu2023convolutions}    & T & CLIP~\cite{radford2021learning} & - & 23.3 \\
      NACLIP \cite{hajimiri2025pay}  & T & CLIP~\cite{radford2021learning} & - & 25.5 \\
      ProxyCLIP \cite{lan2024proxyclip}  & T & CLIP~\cite{radford2021learning} + DINO~\cite{caron2021emerging} & - & 27.1 \\
      Matcher  \cite{liu2023matcher}  & V & DINOv2~\cite{oquab2023dinov2} + SAM~\cite{kirillov2023segment} & 5 imgs/class & {33.5} \\
      GFSAM  \cite{zhang2024bridge}     & V & DINOv2~\cite{oquab2023dinov2} + SAM~\cite{kirillov2023segment} & 5 imgs/class & {38.7} \\

      \midrule
      \multicolumn{5}{l}{\textbf{Few-Shot Visual Adaptation}} \\
      FC-CLIP-L + CoOp \cite{zhou2022learning}             & T+V & CLIP~\cite{radford2021learning} & 5 imgs/class & 20.9{\footnotesize $\pm$0.9} \\
      FC-CLIP-L + CoCoOp \cite{zhou2022conditional}        & T+V & CLIP~\cite{radford2021learning} & 5 imgs/class & 21.4{\footnotesize $\pm$0.8} \\
      FC-CLIP-L + CLIP-Adapter \cite{gao2024clip}          & T+V & CLIP~\cite{radford2021learning} & 5 imgs/class & 26.1{\footnotesize $\pm$0.6} \\
      FC-CLIP-L + TipAdapter-F \cite{zhang2022tip}         & T+V & CLIP~\cite{radford2021learning} & 5 imgs/class & 27.7{\footnotesize $\pm$0.3} \\
      \textbf{\method-L }             & T+V & CLIP~\cite{radford2021learning} & 5 imgs/class & 33.1{\footnotesize $\pm$0.1} \\

      \bottomrule
  \end{tabular}}
  \caption{\textbf{Evaluation on SoT} \cite{rosi2025show}. We carefully implemented the few-shot adaptation baselines while we report official results for the other approaches. For each method we also report the visual encoder(s) employed for a more fair comparison.}
  \label{tab:sot-results}
\end{table}

%% file: tables/ablation_prototypes.tex
\begin{table}[t]
  \centering
  \resizebox{\columnwidth}{!}{
    \begin{tabular}{@{}lccccccc@{}}
      \toprule
      Prototype & \multirow{2}{*}{\shortstack{Trainable\\Parameters}} & \multicolumn{3}{c}{ADE20K} & \multicolumn{3}{c}{Cityscapes} \\
      \cmidrule(lr){3-5} \cmidrule(lr){6-8}
      & & PQ & mIoU & AP & PQ & mIoU & AP \\
      \midrule
      No Protoypes      &  -  & 14.6 & 17.8 & 8.7 & 44.2 & 54.8 & 25.7  \\
      Random            & 39K $|$ 5K   & 26.8 & 33.3 & 16.8 & 48.4 & 64.4 & 28.2  \\
      Class Embedding   & 116K $|$ 15K & 30.1 & 36.6 & 17.4 & 48.4 & 64.6 & \textbf{28.6} \\
      Queries           & 39K $|$ 5K   & 31.7 & 37.7 & 17.8 & 49.3 & 65.1 & 27.4 \\
      Mask pooling      & 39K $|$ 5K   & 28.9 & 36.4 & 16.1 & 49.3 & 64.8 & 27.3 \\
      \textbf{Ours}     & 39K $|$ 5K   & \textbf{32.2} & \textbf{38.7} & \textbf{18.3} & \textbf{50.1} & \textbf{67.7} & 27.9 \\
      \bottomrule
  \end{tabular}}
  \caption{\textbf{Ablation study on different prototype representations.} We compare various prototype initialization strategies and report the number of trainable parameters for each method. The parameter count varies by dataset and is shown as ADE20K $|$ Cityscapes.}
  \label{tab:ablation-prototypes}
\end{table}

%% file: sec/5_conclusion.tex
\section{Conclusion}
\label{sec:conclusion}

We introduce the Few-Shot Visual Adaptation setting for text-prompted Segmentation, identifying misclassification as the primary bottleneck that limits generalization to novel domains. We introduce \textbf{Pr}ototype \textbf{Ada}ptation (\method), a parameter-efficient approach that adapts frozen FC-CLIP models by constructing class-specific prototypes from pixel-level and query embeddings, then fusing visual similarity scores with text-based classification. \method achieves consistent improvements across semantic, instance, and panoptic segmentation on five benchmarks while preserving zero-shot capabilities.

\myparagraph{Broader Impact}
Few-shot adaptation can make segmentation models more accessible in domains with limited data, such as medical or industrial fields. \method lowers annotation requirements, supporting broader deployment even in resource-constrained settings. We hope our work can serve as a reference for future research in this direction.

%% file: sec/X_suppl.tex
\clearpage
\appendix
\normalsize
\setcounter{table}{0}
\setcounter{figure}{0}
\maketitlesupplementary

\appendix
\section*{Appendix}
\label{appendix}
\paragraph{Table of contents:}
\begin{itemize}[itemsep=-1pt,topsep=-1pt]
  \item \S\ref{sec:supp:impl_details}: Implementation Details
  \item \S\ref{sec:supp:ablations}: Additional ablation studies
  \item \S\ref{sec:supp:generalization}: Additional studies on generalization
  \item \S\ref{sec:supp:efficiency}: Efficiency analysis
  \item \S\ref{sec:supp:detailed_results}: Detailed results
  \item \S\ref{sec:supp:qualitative}: Qualitative results
\end{itemize}

\section{Implementation Details}
\label{sec:supp:impl_details}

\subsection{\method}

To train our method we employ AdamW \cite{loshchilov2017decoupled} optimizer with a weight decay of 0.01 with cosine learning rate scheduling. We use a batch size of 8 for all datasets.
We adapt our method to different datasets using a consistent approach while adjusting key hyperparameters based on dataset characteristics. Table~\ref{tab:hyperparameters} summarizes the learning rate, number of iterations, and initial $\alpha$ value used across all datasets. For standard benchmarks (ADE20K \cite{zhou2017scene}, Cityscapes \cite{cordts2016cityscapes}, Mapillary Vistas \cite{neuhold2017mapillary}), we use a learning rate of 0.008 with 1000 iterations and $\alpha_{\text{init}}=80$. The ShowOrTell \cite{rosi2025show} benchmark follows a similar configuration with the same learning rate and $\alpha$ value, but we reduce the number of iterations to 500 for most datasets to balance adaptation efficiency and computational cost. Only UECFOOD \cite{ege2019new}, PASCAL VOC 2012 \cite{everingham2010pascal}, and ZeroWaste \cite{bashkirova2022zerowaste} require 1000 iterations due to their larger domain shift. For the SegInW \cite{zou2023generalized} benchmark, which presents significantly different visual domains, we adopt a more conservative approach using a lower learning rate of 0.0002 with 50 iterations and $\alpha_{\text{init}}=50$. Notably, House-Parts and Strawberry benefit from a higher learning rate (0.002) and more iterations (100), while Trash requires extended adaptation with 800 iterations to handle its challenging characteristics.

\begin{table}[t]
  \setlength{\tabcolsep}{2.5pt}
  \centering
  \small
  \resizebox{\columnwidth}{!}{
    \begin{tabular}{@{}lccc@{}}
      \toprule
      Dataset & LR & Iters & $\alpha_{\text{init}}$ \\
      \midrule
      \multicolumn{4}{l}{\textit{Standard Benchmarks}} \\
      ADE20K \cite{zhou2017scene}, Cityscapes \cite{cordts2016cityscapes}, Mapillary Vistas \cite{neuhold2017mapillary} & 0.008 & 1000 & 80 \\
      \midrule
      \multicolumn{4}{l}{\textit{ShowOrTell Benchmark \cite{rosi2025show}}} \\
      \begin{tabular}[c]{@{}l@{}}House-Parts, LoveDA-Rural \cite{wang2021loveda}, LoveDA-Urban \cite{wang2021loveda},\\MHPv1 \cite{li2017multiple}, PIDray \cite{zhang2023pidray}, Pizza, Toolkits, Trash, UAVid \cite{lyu2020uavid},\\ZeroWaste \cite{bashkirova2022zerowaste}
      \end{tabular} & 0.008 & 500 & 80 \\
      UECFOOD \cite{ege2019new}, PASCAL VOC 2012 \cite{everingham2010pascal} & 0.008 & 1000 & 80 \\
      \midrule
      \multicolumn{4}{l}{\textit{SegInW Benchmark \cite{zou2023generalized}}} \\
      \begin{tabular}[c]{@{}l@{}}Airplane-Parts, Bottles, Brain-Tumor, Chicken,\\Cows, Electric-Shaver, Elephants, Fruits, Garbage,\\Ginger-Garlic, Hand, Hand-Metal, HouseHold-Items,\\Nutterfly-Squireel, Phones, Poles, Puppies, Rail,\\Salmon-Fillet, Tablets, Toolkits, Watermelon
      \end{tabular} & 0.0002 & 50 & 50 \\
      House-Parts, Strawberry & 0.002 & 100 & 50 \\
      Trash & 0.0002 & 800 & 50 \\
      \bottomrule
  \end{tabular}}
  \caption{Hyperparameters used for adaptation across different datasets. We report the learning rate (LR), number of iterations (Iters), and initial $\alpha$ value.}
  \label{tab:hyperparameters}
\end{table}

\subsection{Baselines}
For all baselines, we follow the same training protocol described for our method in the previous section.
For their implementation and choice of hyperparameters, we followed their original publications and officially released code.

\myparagraph{CoOp \cite{zhou2022learning}} For CoOp, we follow the original implementation and we set the context length to 16 with no initialization from hand-crafted prompts or class-specific prompts.

\myparagraph{CoCoOp \cite{zhou2022conditional}} For CoCoOp, we initialize the context vector with ``\textit{This is a photo of a large}'' template and we keep the rest of the hyperparameters as in the original implementation.

\myparagraph{CLIP-Adapter \cite{gao2024clip}} For CLIP-Adapter, we implement the method as described in the original paper, using a 4 times reduction ratio for the adapter's bottleneck layer. We insert the adapter after the MLP that generates the class embeddings.

\myparagraph{TipAdapter \cite{zhang2022tip}} For TipAdapter, we follow the original implementation and we build the cache memory using the masked pooled features from the visual examples. For the scoring function, we set the hyperparameter $\alpha$ to 10.0 and $\beta$ to 1.0 for all the datasets.

\section{Additional ablation studies}
\label{sec:supp:ablations}

\myparagraph{Alpha values} Table~\ref{tab:ablation-alpha} presents an analysis on the parameter $\alpha$, comparing different initialization values under two training strategies: trainable (where $\alpha$ is optimized during adaptation) and fixed (where $\alpha$ remains constant). Our design choice of using a trainable $\alpha$ initialized at 80 achieves the best overall performance, reaching 32.2 PQ and 38.7 mIoU on ADE20K, 50.1 PQ and 67.7 mIoU on Cityscapes, and 33.1 mIoU on ShowOrTell. This configuration outperforms all fixed $\alpha$ variants, demonstrating that allowing $\alpha$ to adapt during training is crucial for balancing the contribution between text-based and visual prototype-based predictions. When $\alpha$ is fixed, performance degrades across all initialization values, with the best fixed configuration ($\alpha=60$) achieving only 31.9 PQ on ADE20K and 33.0 mIoU on ShowOrTell compared to 32.2 PQ and 33.1 mIoU respectively with trainable $\alpha=80$. Interestingly, the results show that our method performs robustly across a broad range of initialization values ($30 \leq \alpha \leq 100$), with trainable $\alpha$ consistently achieving strong performance: values of 30, 60, 80, and 100 yield 30.4, 31.8, 32.2, and 31.7 PQ on ADE20K, and 32.8, 33.1, 33.1, and 33.1 mIoU on ShowOrTell respectively. However, initializing with very low values (\eg, $\alpha=10$) yields suboptimal performance (27.0 PQ on ADE20K and 31.0 mIoU on ShowOrTell), as the model struggles to properly weight the visual prototypes. This demonstrates that while training $\alpha$ is essential for optimal performance, the method is relatively insensitive to the specific initialization choice within a reasonable range.

\input{tables/supp_ablation_alpha.tex}

\begin{figure}[t]
  \centering
  \includegraphics[width=1\linewidth]{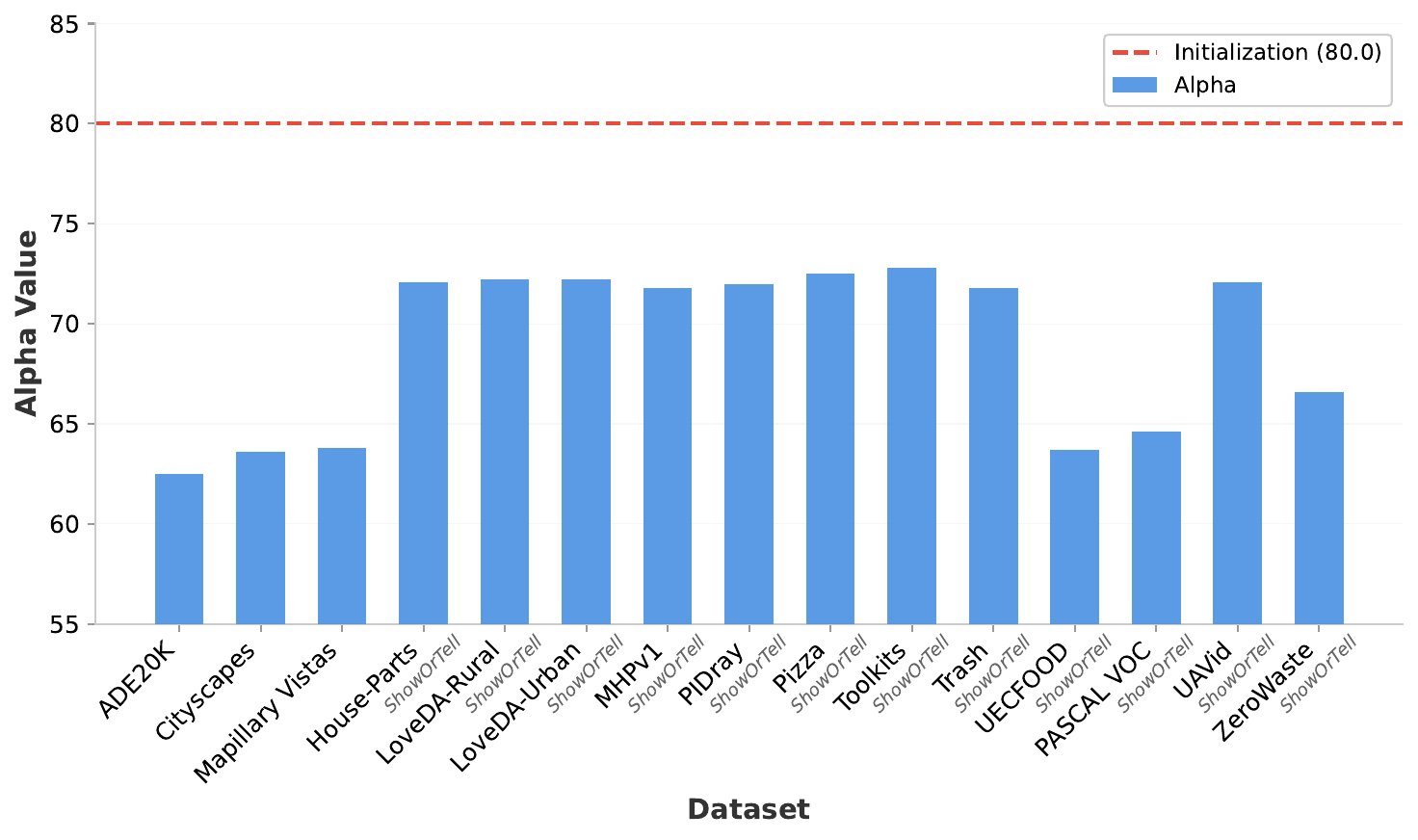}
  \caption{\textbf{Final $\alpha$ values across datasets.} We report the learned $\alpha$ values after training convergence for standard benchmarks (ADE20K, Cityscapes, Mapillary Vistas) and individual datasets in the ShowOrTell \cite{rosi2025show} benchmark. The red dashed line indicates the initialization value ($\alpha_{\text{init}}=80$).}
  \label{fig:supp:alpha}
\end{figure}

\myparagraph{Alpha after training} Table~\ref{tab:alpha-evolution} and Figure~\ref{fig:supp:alpha} show the evolution of the $\alpha$ parameter during training across different benchmarks and individual datasets. We report both the initial value $\alpha_{\text{init}}$ used to start the adaptation process and the final value $\alpha_{\text{final}}$ after training converges. The results reveal that $\alpha$ adapts differently across benchmarks based on their specific characteristics. For standard benchmarks (ADE20K, Cityscapes, Mapillary Vistas), $\alpha$ consistently decreases from its initial value of 80, converging to values around 63 (62.5, 63.6, and 63.8 respectively). This reduction of approximately 20\% suggests that these well-aligned datasets benefit from a more balanced combination of text-based and visual prototype-based predictions, with the model learning to place slightly less emphasis on visual prototypes while still maintaining their contribution. In contrast, for the SegInW \cite{zou2023generalized} benchmark, where we initialize $\alpha$ at 50 due to the more diverse visual domains, the final value remains stable at 50.0. This stability is primarily due to the limited number of training iterations and the relatively small number of classes in many SegInW datasets, which constraints the extent to which $\alpha$ can be effectively optimized during the short adaptation phase. The ShowOrTell \cite{rosi2025show} benchmark shows an intermediate behavior, with $\alpha$ decreasing from 80 to 70.4 on average, a more moderate reduction compared to standard benchmarks. As illustrated in Figure~\ref{fig:supp:alpha}, the final $\alpha$ values across individual ShowOrTell datasets exhibit notable variation, ranging from approximately 64 (PASCAL VOC, UECFOOD, ZeroWaste) to 73 (House-Parts, Toolkits, Pizza). This suggests that ShowOrTell's specialized domains (\eg, aerial imagery, X-ray scans, waste management) require stronger visual prototype influence than standard scene parsing datasets but still benefit from adaptation. These adaptive behaviors demonstrate that allowing $\alpha$ to be trainable enables the model to automatically discover the optimal balance between text and visual prototypes for each specific domain.

\begin{table}[t]
  \centering
  \small
  \resizebox{\columnwidth}{!}{
    \begin{tabular}{@{}lccccc@{}}
      \toprule
      & ADE20K & Cityscapes & Mapillary & SegInW & ShowOrTell \\
      \midrule
      $\alpha_{\text{init}}$ & 80 & 80 & 80 & 50 & 80 \\
      $\alpha_{\text{final}}$ & 62.5{\footnotesize $\pm$0.0} & 63.6{\footnotesize $\pm$0.1} & 63.8{\footnotesize $\pm$0.0} & 50.0{\footnotesize $\pm$0.0} & 70.4{\footnotesize $\pm$3.3} \\
      \bottomrule
  \end{tabular}}
  \caption{\textbf{Evolution of $\alpha$ parameter during training.} We report the initial value ($\alpha_{\text{init}}$) and the final value after training convergence ($\alpha_{\text{final}}$) across different benchmarks.}
  \label{tab:alpha-evolution}
\end{table}

\begin{table}[t]
  \centering
  \resizebox{\columnwidth}{!}{
    \begin{tabular}{@{}c@{\hspace{6pt}}c@{\hspace{6pt}}c@{\hspace{6pt}}c@{\hspace{6pt}}c@{\hspace{8pt}}c@{\hspace{8pt}}c@{\hspace{6pt}}c@{}}
      \toprule
      \multirow{2}{*}{imgs/class} & \multirow{2}{*}{Prompt Type} & \multicolumn{3}{c}{ADE20K} & \multicolumn{3}{c}{Cityscapes} \\
      \cmidrule(lr){3-5} \cmidrule(lr){6-8}
      & & PQ & mIoU & AP & PQ & mIoU & AP \\
      \midrule
      1 & V+T   & 25.1{\footnotesize $\pm$1.1} & 30.2{\footnotesize $\pm$1.0} & 16.3{\footnotesize $\pm$0.9} & 45.6{\footnotesize $\pm$1.4} & 60.1{\footnotesize $\pm$2.6} & 23.0{\footnotesize $\pm$2.5} \\
      2 & V+T   & 28.4{\footnotesize $\pm$0.4} & 34.9{\footnotesize $\pm$0.8} & 17.0{\footnotesize $\pm$0.3} & 48.4{\footnotesize $\pm$0.4} & 63.2{\footnotesize $\pm$1.9} & 26.3{\footnotesize $\pm$1.1} \\
      5 & V only & 29.3{\footnotesize $\pm$0.3} & 36.6{\footnotesize $\pm$0.5} & 16.1{\footnotesize $\pm$0.4} & 47.7{\footnotesize $\pm$0.8} & 62.8{\footnotesize $\pm$1.0} & 23.2{\footnotesize $\pm$0.9} \\
      \midrule
      5 & V+T & \textbf{31.4}{\footnotesize $\pm$0.7} & \textbf{38.2}{\footnotesize $\pm$0.5} & \textbf{18.1}{\footnotesize $\pm$0.4} & \textbf{49.8}{\footnotesize $\pm$0.5} & \textbf{66.2}{\footnotesize $\pm$1.5} & \textbf{27.9}{\footnotesize $\pm$0.7} \\
      \bottomrule
  \end{tabular}}
  \caption{\textbf{Ablation on prompt type and number of support images.} We compare visual-only (V) and joint visual-textual (V+T) prompting strategies across varying numbers of support images per class on ADE20K \cite{zhou2017scene} and Cityscapes \cite{cordts2016cityscapes}.}
  \label{tab:results}
\end{table}

\begin{table}[t]
  \centering
  \renewcommand{\arraystretch}{0.9}
  \resizebox{\columnwidth}{!}{
    \begin{tabular}{@{}l@{\hspace{6pt}}c@{\hspace{6pt}}c@{\hspace{6pt}}c@{\hspace{8pt}}c@{\hspace{8pt}}c@{\hspace{6pt}}c@{}}
      \toprule
      \multirow{2}{*}{Model} & \multicolumn{3}{c}{ADE20K} & \multicolumn{3}{c}{Cityscapes} \\
      \cmidrule(lr){2-4} \cmidrule(lr){5-7}
      & PQ & mIoU & AP & PQ & mIoU & AP \\
      \midrule
      MAFT$+$ \cite{jiao2024collaborative}  & 27.1 & 33.9 & 15.7 & 38.3 & 52.8 & 20.3 \\
      \textbf{\method-L} (MAFT$+$)   & 28.0{\footnotesize $\pm$0.1} & 34.9{\footnotesize $\pm$0.1} & 16.0{\footnotesize $\pm$0.1} & 41.3{\footnotesize $\pm$0.2} & 55.0{\footnotesize $\pm$0.3} & 21.0{\footnotesize $\pm$0.3} \\
      \bottomrule
  \end{tabular}}
  \caption{\textbf{Application of \method to MAFT+ \cite{jiao2024collaborative}}. We report the performance of MAFT+ and our method when applied to MAFT+ across ADE20K \cite{zhou2017scene} and Cityscapes \cite{cordts2016cityscapes}.}
  \label{tab:maft_results}
\end{table}

\myparagraph{Prompt type and number of support images} Table~\ref{tab:results} ablates two key factors of our adaptation pipeline: the type of prompt used and the number of support images per class. Row~(A) represents our full method using 5 images per class with combined visual and textual prompts (V+T), achieving the best performance on both ADE20K (31.4 PQ, 38.2 mIoU) and Cityscapes (49.8 PQ, 66.2 mIoU). Comparing rows~(A) and~(B) isolates the contribution of textual prompts: removing text and relying solely on visual prototypes (V only) consistently degrades performance by 2.1 PQ and 1.6 mIoU on ADE20K, confirming that combining visual and textual cues is beneficial. Rows~(C) and~(D) study the effect of reducing the number of support images to 1 and 2 respectively. Performance drops substantially with a single support image (row~C), yielding 25.1 PQ and 30.2 mIoU on ADE20K, while using 2 images (row~D) recovers most of the gap (28.4 PQ). These results show that our method scales gracefully with the number of support images, but benefits most from having at least 5 images per class to build reliable visual prototypes.

\myparagraph{Application to other methods} \method can be applied to any open-vocabulary segmentation model built on M2F, including more recent methods like MAFT+ \cite{jiao2024collaborative}. As depicted in Table~\ref{tab:maft_results}, applying \method to MAFT+ yields consistent improvements across both ADE20K and Cityscapes, demonstrating the generality of our visual prototype learning strategy. On ADE20K, \method-L (MAFT+) achieves 28.0 PQ and 34.9 mIoU, improving over the original MAFT+ by 0.9 PQ and 1.0 mIoU. On Cityscapes, \method-L (MAFT+) reaches 41.3 PQ and 55.0 mIoU, outperforming MAFT+ by 3.0 PQ and 2.2 mIoU. These results confirm that our method can effectively enhance the adaptation capabilities of various open-vocabulary segmentation models by learning visual prototypes that complement their existing architectures.

\section{Additional studies on generalization}
\label{sec:supp:generalization}

\begin{figure*}[t]
  \centering
  \includegraphics[width=1\linewidth]{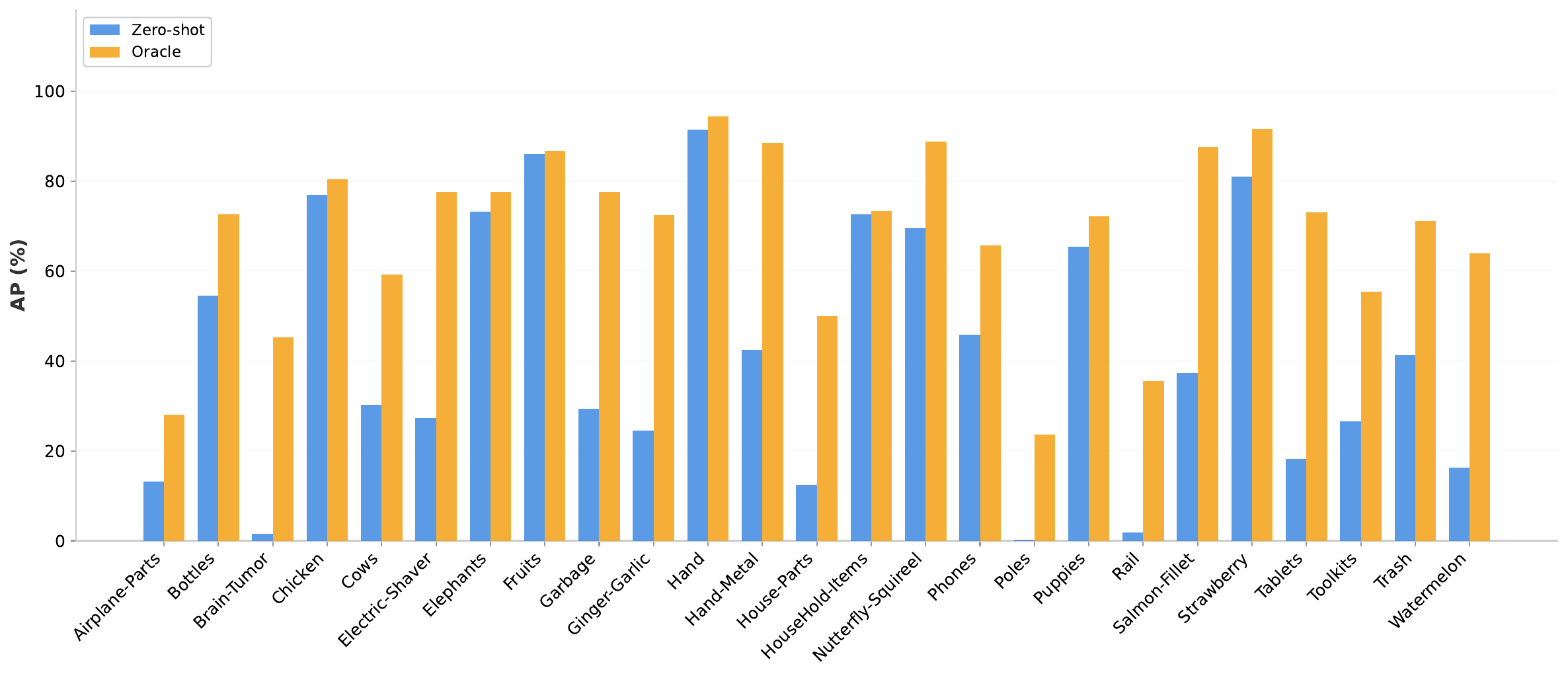}
  \caption{\textbf{Complete oracle results on SegInW \cite{zou2023generalized} datasets.} We report zero-shot performance (mAP) of the original FC-CLIP \cite{yu2023convolutions} model across all 25 datasets in the SegInW benchmark, evaluated only on the classes present in the ground-truth masks. While some datasets show zero-shot performance close to the oracle upper bound, indicating strong pre-training alignment with those visual domains, other datasets exhibit significantly lower performance, revealing substantial domain gaps that highlight the need for adaptation.}
  \label{fig:supp:oracle_seginw}
\end{figure*}

We extend the evaluation of zero-shot capabilities of FC-CLIP \cite{yu2023convolutions} on the SegInW benchmark presented in the main paper by reporting a complete comparison between zero-shot and oracle performance across all 25 datasets.

\myparagraph{Oracle analysis on SegInW} Figure~\ref{fig:supp:oracle_seginw} presents a comprehensive comparison between zero-shot and oracle performance across all 25 datasets in the SegInW \cite{zou2023generalized} benchmark. The oracle setting represents an upper bound where the model is evaluated using the classes present in the ground-truth masks, focusing purely on the model's ability to recognize and segment specific object categories. The results reveal significant heterogeneity in the zero-shot capabilities of FC-CLIP \cite{yu2023convolutions} across different domains. On one end of the spectrum, datasets like Hand, Chicken, and Fruits show relatively small gaps between zero-shot and oracle performance (less than 10 mAP difference), indicating that CLIP's pre-training already provides strong visual-semantic alignment for these common object categories. These datasets benefit from rich representation in web-scale training data, making few-shot adaptation less critical. Conversely, datasets such as Salmon-Fillet, Brain-Tumor, Electric-Shaver, and Watermelon exhibit substantial performance gaps exceeding 40-50 mAP, revealing severe domain misalignment. These specialized domains, featuring technical components, infrastructure elements, or objects with high visual ambiguity, are poorly represented in CLIP's pre-training data, making them prime candidates for few-shot adaptation. Datasets like Poles, Rail, and Cows show intermediate gaps (20-30 mAP), where zero-shot performance is moderate but substantial improvement is possible through adaptation.

\section{Efficiency Analysis}
\label{sec:supp:efficiency}
To better asses the efficiency of our method, we analyze the number of trainable parameters and training time required for adaptation across different datasets. Since we are training only the visual prototypes and the alpha parameter, the number of trainable parameters can be calculated as follows:
\begin{equation}
  P_{\text{trainable}} = (N_{\text{classes}} + 1) \times D_{\text{}} + 1
\end{equation}
where $(N_{\text{classes}} + 1)$ is the number of classes in the dataset plus the \textit{void} class, $D_{\text{}}$ is the dimensionality of the feature space (\ie in our case $256$).

For instance, ADE20K \cite{zhou2017scene} with 150 classes requires 38,657 trainable parameters, Mapillary Vistas \cite{neuhold2017mapillary} with 65 classes requires 16,897 parameters, and Cityscapes \cite{cordts2016cityscapes} with 19 classes requires only 5,121 parameters.
Thanks to this minimal parameter footprint, our method enables rapid adaptation. Adapting to standard benchmarks such as ADE20K or Cityscapes takes less than \textit{30 minutes} on a single NVIDIA A5000 GPU, making our approach practical for real-world scenarios requiring efficient domain adaptation.

\section{Detailed results}
\label{sec:supp:detailed_results}

In the following sections, we provide comprehensive results for our method and all baselines across the ShowOrTell \cite{rosi2025show} and SegInW \cite{zou2023generalized} benchmarks. We analyze performance on each individual dataset, highlighting strengths and weaknesses of our approach compared to few-shot adaptation baselines.

\subsection{Detailed results on SegInW}

Table~\ref{tab:supp:seginw_results} presents detailed results for our method and all baselines across the 25 diverse datasets in the SegInW \cite{zou2023generalized} benchmark. Our method achieves the best average performance (43.3 mAP), outperforming FC-CLIP (41.6 mAP), CLIP-Adapter (42.1 mAP), CoOp (41.2 mAP), and CoCoOp (41.5 mAP). While no single method dominates across all datasets, our approach demonstrates particular strength in challenging scenarios with complex object categories, achieving the best performance on 8 datasets including Brain-Tumor (+3.0 mAP over FC-CLIP), Chicken (+0.8 mAP), Cows (+2.4 mAP), Phones (+3.6 mAP), Puppies (+5.0 mAP), Tablets (+24.0 mAP), and Toolkits (+8.9 mAP). Notably, our method shows robust performance across the highly varied domains in SegInW \cite{zou2023generalized}, from medical imaging (Brain-Tumor) to animals (Chicken, Cows, Puppies) to everyday objects (Phones, Tablets), highlighting the effectiveness of our visual prototype learning strategy for cross-domain adaptation. However, our method underperforms on certain datasets where the baseline FC-CLIP excels, particularly Bottles (32.8 mAP vs. 54.6 mAP), Fruits (47.4 mAP vs. 86.0 mAP), and Trash (23.3 mAP vs. 41.3 mAP). These cases suggest that when the frozen text embeddings already provide strong semantic alignment with the visual domain, our prototype adaptation may introduce unnecessary complexity. Additionally, datasets like Elephants (70.2 mAP vs. 73.3 mAP for FC-CLIP) and HouseHold-Items (61.4 mAP vs. 72.7 mAP) show that our method can struggle when few-shot examples are insufficient to capture the high intra-class variability present in these categories.

\subsection{Detailed results on ShowOrTell}

Table~\ref{tab:supp:showortell_results} reports comprehensive results across the 14 datasets in the ShowOrTell \cite{rosi2025show} benchmark, including 12 target datasets plus ADE20K and Cityscapes. Our method achieves the highest average performance (33.1 mIoU), significantly outperforming the second-best TipAdapter-F (27.7 mIoU) by 5.4 mIoU. The improvements are particularly pronounced on datasets with large domain shifts from natural images: House-Parts (30.8 mIoU vs. 15.0 mIoU for TipAdapter-F), Pizza (30.7 mIoU vs. 19.5 mIoU), Zero-Waste (28.5 mIoU vs. 14.3 mIoU), and PASCAL VOC (74.5 mIoU vs. 67.8 mIoU). Our method achieves the best performance on 8 out of 14 datasets, demonstrating consistent superiority over prompt-learning methods (CoOp, CoCoOp) and adapter-based approaches (CLIP-Adapter, TipAdapter-F). The results confirm that learning visual prototypes in the feature space provides more robust adaptation than text-based prompt tuning, especially when dealing with specialized visual domains like aerial imagery (UAVid), waste management (Zero-Waste, Trash), and food recognition (Pizza, UECFood). Nonetheless, our method shows weaker performance on certain outdoor scene parsing datasets where TipAdapter-F performs better, specifically LoveDA-Rural (25.5 mIoU vs. 30.9 mIoU) and LoveDA-Urban (30.6 mIoU vs. 40.1 mIoU). Similarly, on UECFood (20.4 mIoU vs. 21.9 mIoU for TipAdapter-F) and Cityscapes (66.2 mIoU vs. 64.3 mIoU for CLIP-Adapter), the performance gaps are minimal, suggesting that certain well-structured domains with consistent visual appearance may not fully benefit from our prototype learning approach.

\input{tables/supp_seginw_results.tex}
\input{tables/supp_showortell_results.tex}

\section{Qualitative results}
\label{sec:supp:qualitative}

\begin{figure*}[t]
  \centering
  \includegraphics[width=0.85\linewidth]{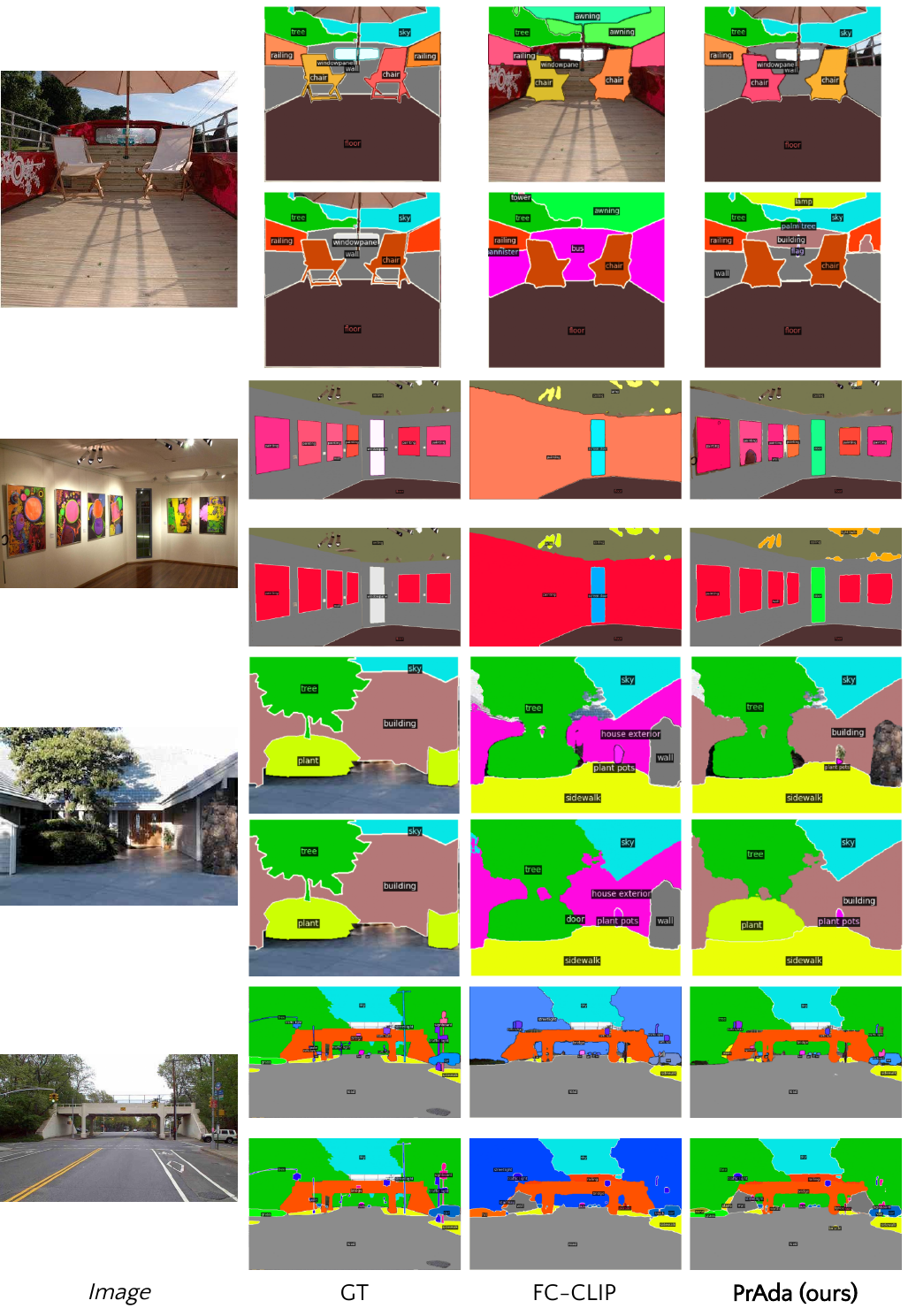}
  \caption{\textbf{Qualitative results on ADE20K \cite{zhou2017scene}.} For each image, the first row contains \textit{panoptic segmentation} predictions, while the second row contains \textit{semantic segmentation} predictions.}
  \label{fig:supp:qualitative_ade}
\end{figure*}

\begin{figure*}[t]
  \centering
  \includegraphics[width=0.9\linewidth]{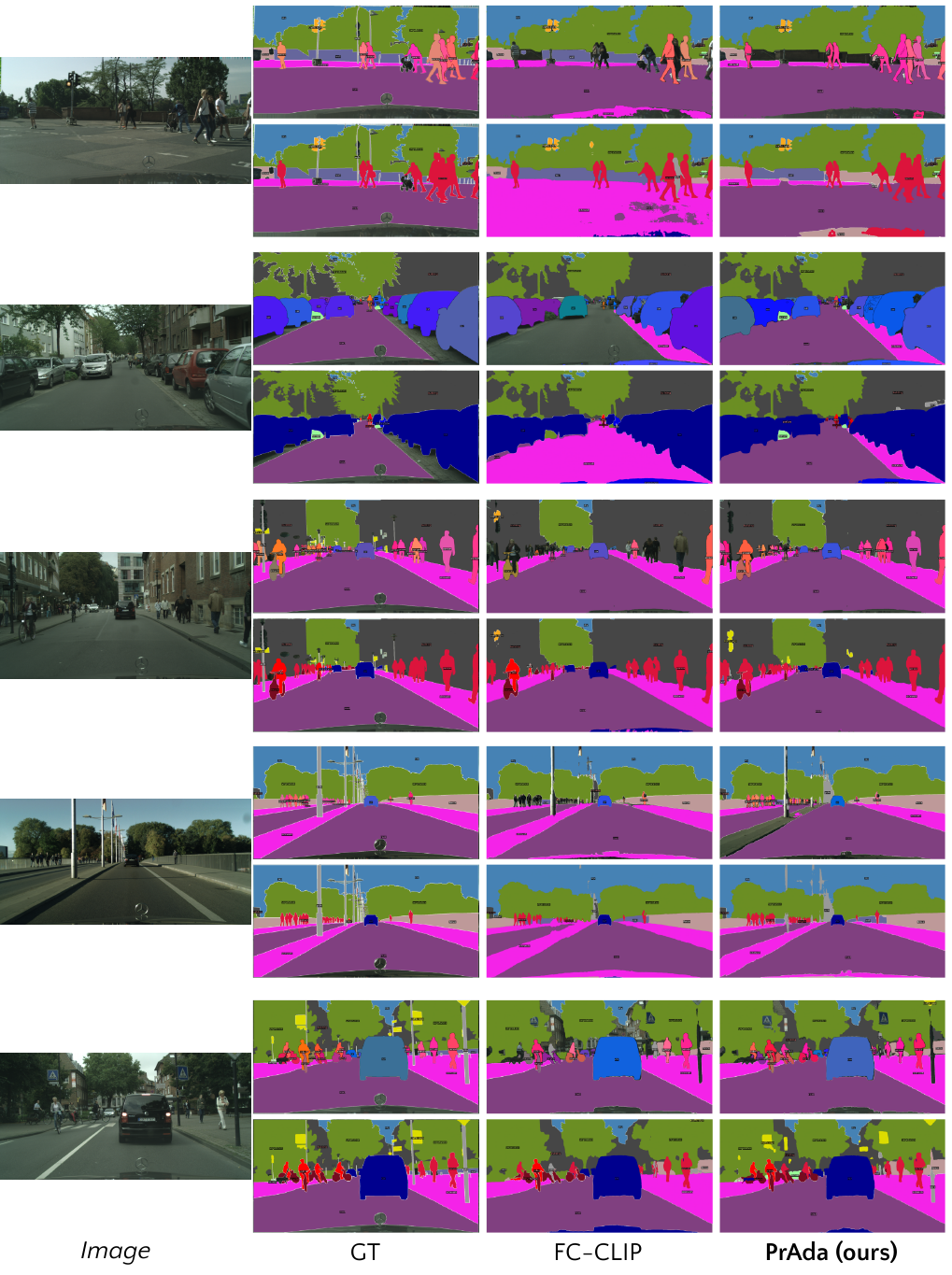}
  \caption{\textbf{Qualitative results on Cityscapes \cite{cordts2016cityscapes}.} For each image, the first row contains \textit{panoptic segmentation} predictions, while the second row contains \textit{semantic segmentation} predictions.}
  \label{fig:supp:qualitative_city}
\end{figure*}

\begin{figure*}[t]
  \centering
  \includegraphics[width=0.8\linewidth]{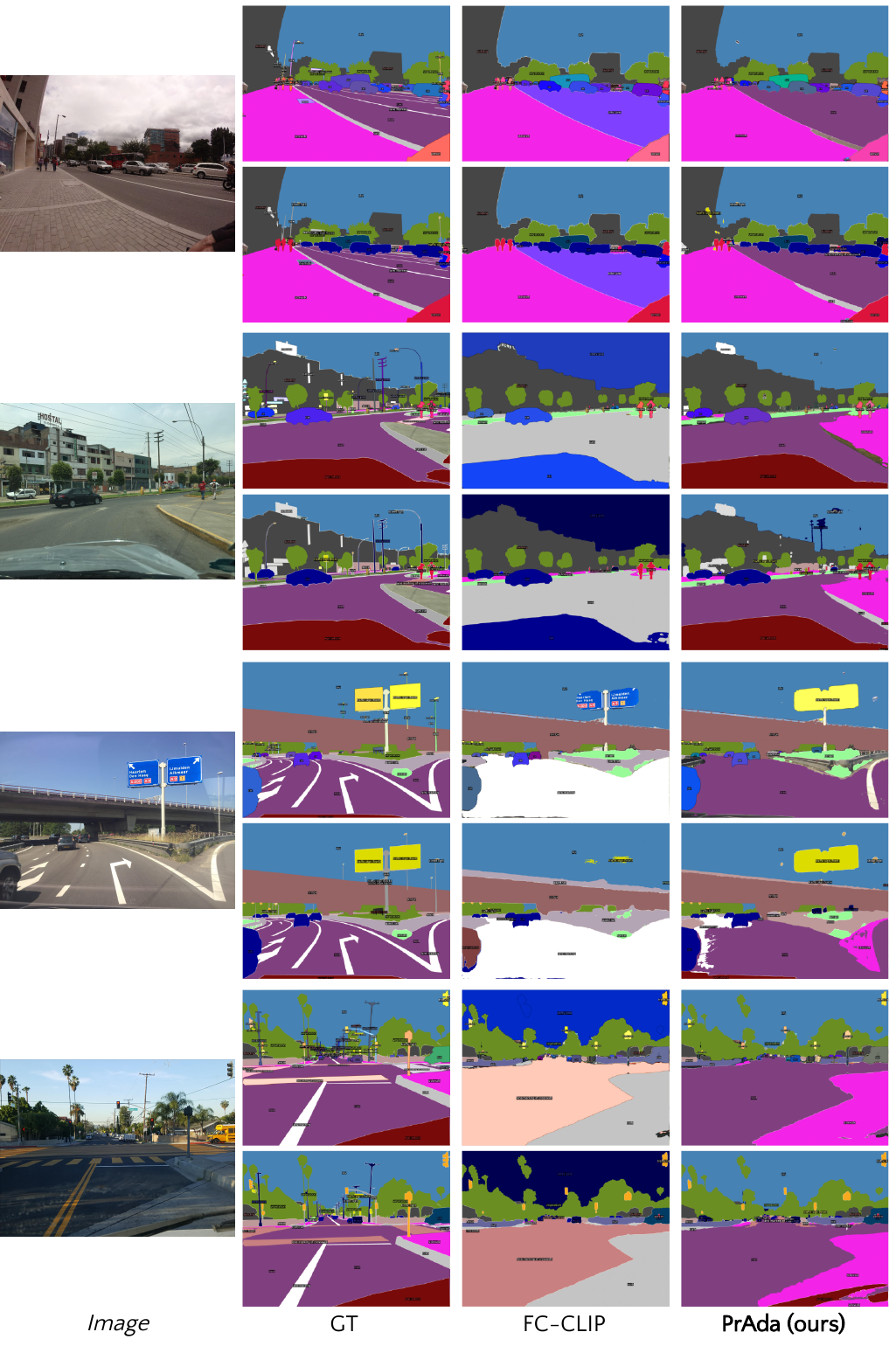}
  \caption{\textbf{Qualitative results on Mapillary Vistas \cite{neuhold2017mapillary}.} For each image, the first row contains \textit{panoptic segmentation} predictions, while the second row contains \textit{semantic segmentation} predictions.}
  \label{fig:supp:qualitative_mapillary}
\end{figure*}

\begin{figure*}[t]
  \centering
  \includegraphics[width=0.75\linewidth]{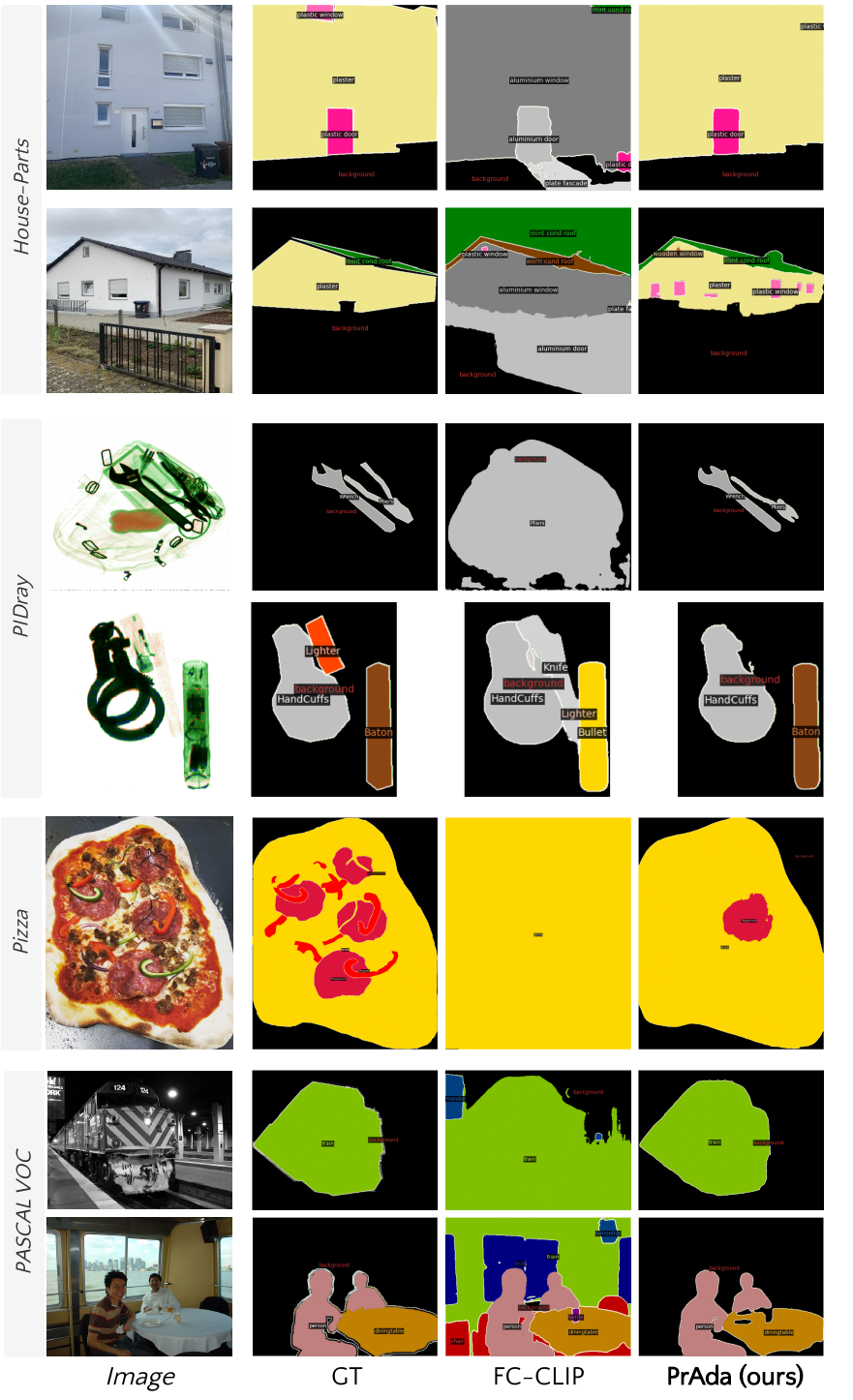}
  \caption{\textbf{Qualitative results on ShowOrTell \cite{rosi2025show} datasets.} We select a subset of datasets composed by: House-Parts, PIDray, Pizza and PASCAL VOC.}
  \label{fig:supp:qualitative_sot}
\end{figure*}

In this section, we present qualitative results to complement the quantitative analysis provided in the main paper and previous sections. We visualize predictions from our method across different benchmarks: ADE20K \cite{zhou2017scene} (\cref{fig:supp:qualitative_ade}), Cityscapes \cite{cordts2016cityscapes} (\cref{fig:supp:qualitative_city}), Mapillary Vistas \cite{neuhold2017mapillary} (\cref{fig:supp:qualitative_mapillary}), and diverse domains from the ShowOrTell \cite{rosi2025show} benchmark (\cref{fig:supp:qualitative_sot}). These visualizations illustrate how our visual prototype learning approach adapts to varying visual domains and demonstrate the quality of both panoptic and semantic segmentation predictions. For each example, we show the input image alongside the predicted masks, highlighting the model's ability to accurately segment objects and stuff categories across different scenarios.

%% file: tables/supp_ablation_alpha.tex
\begin{table}[t]
  \centering
  \resizebox{\columnwidth}{!}{
    \begin{tabular}{@{}lcccccccc@{}}
      \toprule
      \multirow{2}{*}{$\alpha$} & \multirow{2}{*}{Trainable} & \multicolumn{3}{c}{ADE20K} & \multicolumn{2}{c}{Cityscapes} & SoT \\
      \cmidrule(lr){3-5} \cmidrule(lr){6-7} \cmidrule(lr){8-8}
      & & PQ & mIoU & AP & PQ & mIoU & mIoU \\
      \midrule
      \multicolumn{8}{c}{\textit{Trainable $\alpha$}} \\
      \midrule
      10  & \checkmark & 27.0 & 35.5 & 18.4 & 48.6 & 65.9 & 31.0 \\
      30  & \checkmark & 30.4 & 37.9 & 18.8 & 49.8 & 67.0 & 32.8 \\
      60  & \checkmark & 31.8 & 39.1 & 18.4 & 49.3 & 67.0 & 33.1 \\
      \textbf{80}  & \checkmark & 32.2 & 38.7 & 18.3 & 50.1 & 67.7 & 33.1 \\
      100 & \checkmark & 31.7 & 38.2 & 17.8 & 49.5 & 67.1 & 33.1 \\
      \midrule
      \multicolumn{8}{c}{\textit{Fixed $\alpha$}} \\
      \midrule
      10  & \xmark & 26.9 & 35.2 & 18.3 & 48.5 & 65.7 & 30.6 \\
      30  & \xmark & 30.8 & 38.3 & 18.6 & 49.6 & 66.9 & 32.9 \\
      60  & \xmark & 31.9 & 38.8 & 18.0 & 49.1 & 66.0 & 33.0 \\
      80  & \xmark & 31.6 & 38.0 & 17.6 & 49.3 & 67.2 & 33.0 \\
      100 & \xmark & 31.2 & 37.5 & 17.1 & 48.8 & 65.9 & 32.7 \\
      \bottomrule
  \end{tabular}}
  \caption{\textbf{Ablation study on different values of $\alpha$.} We compare different initialization values for $\alpha$ parameter with two training strategies: trainable (\ie $\alpha$ is learned during training) and fixed (\ie $\alpha$ remains constant).}
  \label{tab:ablation-alpha}
\end{table}

%% file: tables/supp_seginw_results.tex
\begin{table*}[t]
  \setlength{\tabcolsep}{1pt}
  \centering
  \scriptsize
  \resizebox{\textwidth}{!}{
    \begin{tabular}{@{}lccccccccccccccccccccccccccc@{}}
      \toprule
      Method &
      \begin{tabular}[c]{@{}c@{}}Airplane-\\Parts
      \end{tabular} & Bottles &
      \begin{tabular}[c]{@{}c@{}}Brain-\\Tumor
      \end{tabular} & Chicken & Cows &
      \begin{tabular}[c]{@{}c@{}}Electric-\\Shaver
      \end{tabular} & Elephants & Fruits & Garbage &
      \begin{tabular}[c]{@{}c@{}}Ginger-\\Garlic
      \end{tabular} & Hand &
      \begin{tabular}[c]{@{}c@{}}Hand-\\Metal
      \end{tabular} &
      \begin{tabular}[c]{@{}c@{}}House-\\Parts
      \end{tabular} &
      \begin{tabular}[c]{@{}c@{}}HouseHold-\\Items
      \end{tabular} &
      \begin{tabular}[c]{@{}c@{}}Nutterfly-\\Squireel
      \end{tabular} & Phones & Poles & Puppies & Rail &
      \begin{tabular}[c]{@{}c@{}}Salmon-\\Fillet
      \end{tabular} & Strawberry & Tablets & Toolkits & Trash & Watermelon & \textbf{AVG} \\
      \midrule
      \textit{FC-CLIP} & \underline{13.2} & \textbf{54.6} & \underline{1.6} & 76.9 & 30.2 & 27.3 & 73.3 & \textbf{86.0} & 29.4 & \underline{24.5} & \underline{91.5} & 42.5 & \textbf{12.4} & \textbf{72.7} & 69.5 & 45.9 & 0.2 & 65.5 & 1.8 & 37.3 & \textbf{81.1} & 18.2 & 26.6 & \textbf{41.3} & 16.2 & 41.6 \\
      \midrule
      FC-CLIP + CoOp & 12.6 & 24.3 & 0.1 & 68.1 & 18.8 & 55.5 & \textbf{74.8} & \underline{85.3} & \underline{35.8} & 21.7 & \textbf{94.0} & \textbf{61.2} & 7.3 & \textbf{72.7} & 70.9 & 47.1 & 1.2 & 62.8 & 1.6 & 26.7 & 78.1 & 15.8 & 31.5 & 35.8 & 26.8 & 41.2 \\
      FC-CLIP + CoCoOp & 12.9 & 19.7 & 0.3 & \underline{78.2} & 29.8 & \textbf{73.4} & \textbf{74.8} & 79.0 & 21.7 & 30.9 & \textbf{94.0} & \underline{55.0} & 9.0 & \textbf{72.7} & \textbf{72.7} & 48.5 & \textbf{2.7} & 65.5 & 2.1 & 31.6 & 78.5 & 21.1 & 15.5 & 32.2 & 14.6 & 41.5 \\
      FC-CLIP + CLIP-Adapter & 12.1 & \underline{51.6} & 0.6 & 74.8 & \underline{35.9} & 70.6 & \underline{74.1} & 55.7 & \textbf{40.4} & 23.4 & \textbf{94.0} & 50.0 & 7.6 & \underline{62.7} & \underline{71.8} & 49.0 & 0.5 & 62.9 & 1.2 & \textbf{37.3} & 60.6 & 20.9 & 28.8 & 39.5 & 25.8 & \underline{42.1} \\
      \midrule
      \textbf{\method} & \textbf{13.5} & 32.8 & \textbf{4.6} & \textbf{79.0} & \textbf{38.3} & \underline{73.1} & 70.2 & 47.4 & 34.0 & \textbf{29.2} & \textbf{94.0} & 54.0 & \underline{10.3} & 61.4 & 71.7 & \textbf{53.6} & 1.6 & \textbf{70.5} & \textbf{2.3} & 22.5 & 75.1 & \textbf{44.2} & \textbf{35.5} & 23.3 & \textbf{40.7} & \textbf{43.3} \\
      \bottomrule
  \end{tabular}}
  \caption{\textbf{Results for few-shot adaptation methods across 25 SegInW datasets.} For each dataset we report the average across 5 random seeds. We report mean Average Precision (mAP).}
  \label{tab:supp:seginw_results}
\end{table*}

%% file: tables/supp_showortell_results.tex
\begin{table*}[t]
  \setlength{\tabcolsep}{2pt}
  \centering
  \scriptsize
  \resizebox{\textwidth}{!}{
    \begin{tabular}{@{}lcccccccccccccccc@{}}
      \toprule
      Method &
      \begin{tabular}[c]{@{}c@{}}House-\\Parts
      \end{tabular} &
      \begin{tabular}[c]{@{}c@{}}LoveDA-\\Rural
      \end{tabular} &
      \begin{tabular}[c]{@{}c@{}}LoveDA-\\Urban
      \end{tabular} & MHPv1 & PIDray & Pizza & Toolkits & Trash & UECFood &
      \begin{tabular}[c]{@{}c@{}}PASCAL\\VOC
      \end{tabular} & UAVid &
      \begin{tabular}[c]{@{}c@{}}Zero-\\Waste
      \end{tabular} & ADE20K &
      \begin{tabular}[c]{@{}c@{}}City-\\scapes
      \end{tabular} & \textbf{AVG} \\
      \midrule
      \textit{FC-CLIP} & 7.5 & \underline{29.3} & \underline{39.1} & 8.0 & 8.9 & 15.5 & 20.1 & 10.5 & 19.4 & 38.1 & 33.3 & 5.5 & 34.1 & 56.2 & 23.3 \\
      \midrule
      FC-CLIP + CoOp & 12.3 & 27.0 & 32.5 & 2.6 & 10.7 & 13.2 & 7.9 & 12.7 & 8.9 & 39.0 & 30.0 & 5.0 & 31.6 & 59.3 & 20.9 \\
      FC-CLIP + CoCoOp & 10.3 & 25.7 & 32.5 & 3.5 & 8.3 & 12.9 & 6.8 & 10.4 & 12.6 & 41.9 & 33.6 & 9.3 & 31.9 & 60.5 & 21.4 \\
      FC-CLIP + CLIP-Adapter & 9.9 & 28.7 & 34.8 & 8.9 & 11.3 & 16.3 & \textbf{27.9} & \underline{15.8} & \underline{21.1} & 46.5 & 35.7 & 6.3 & 37.5 & \textbf{64.3} & 26.1 \\
      FC-CLIP + TipAdapter-F & \underline{15.0} & \textbf{30.9} & \textbf{40.1} & \underline{11.4} & \underline{18.1} & \underline{19.5} & 11.6 & 13.5 & \textbf{21.9} & \underline{67.8} & \underline{37.7} & \underline{14.3} & \textbf{38.5} & 48.0 & \underline{27.7} \\
      \midrule
      \textbf{\method} & \textbf{30.8} & 25.5 & 30.6 & \textbf{12.6} & \textbf{19.6} & \textbf{30.7} & \underline{27.3} & \textbf{17.3} & 20.4 & \textbf{74.5} & \textbf{41.3} & \textbf{28.5} & \underline{38.2} & \textbf{66.2} & \textbf{33.1} \\
      \bottomrule
  \end{tabular}}
  \caption{Results for few-shot adaptation methods across ShowOrTell datasets. For each dataset we report the average across 5 random seeds. We report mean Intersection over Union (mIoU).}
  \label{tab:supp:showortell_results}
\end{table*}